\documentclass{article}

\usepackage{amsmath,amsfonts,bm}

\def\eqref#1{equation~\ref{#1}}

\def\1{\bm{1}}

\DeclareMathAlphabet{\mathsfit}{\encodingdefault}{\sfdefault}{m}{sl}
\SetMathAlphabet{\mathsfit}{bold}{\encodingdefault}{\sfdefault}{bx}{n}

\usepackage{xcolor}
\definecolor{mycitecolor}{RGB}{102, 102, 102}
\usepackage[hidelinks,colorlinks, linkcolor=red, citecolor={mycitecolor}]{hyperref}
\usepackage{url}
\usepackage{amsmath}
\usepackage{algorithm}
\usepackage{algorithmic}
\usepackage{multirow}
\usepackage{verbatim}
\usepackage{gensymb}  
\usepackage{subfiles} 
\usepackage{makecell}
\usepackage{amssymb}  
\usepackage{pifont}
\usepackage{graphicx}
\usepackage{subcaption}
\usepackage{booktabs}
\usepackage{tablefootnote}
\usepackage[flushleft]{threeparttable}

\usepackage{lipsum}
\usepackage{bbding}
\usepackage{enumitem}
\usepackage{ulem}
\usepackage{tcolorbox}       
\usepackage{amsfonts}       
\usepackage{nicefrac}       
\usepackage{microtype}      
\usepackage[table]{xcolor}
\usepackage{xspace}
\usepackage{wrapfig}

\definecolor{mygreen}{HTML}{3cb44b}

\newcommand{\toolname}{UGround\xspace}
\newcommand{\eg}{\textit{e.g.}\xspace}

\newcommand{\ie}{\textit{i.e.}\xspace}

\newcommand{\wrt}{w.r.t.\xspace}

\definecolor{mygray}{gray}{.9}
\definecolor{colorlink}{RGB}{237,2,140}
\usepackage{microtype}
\usepackage{graphicx}
\usepackage{subcaption}
\usepackage{booktabs} 

\usepackage{hyperref}

\usepackage[accepted]{icml2026}

\usepackage{amsmath}
\usepackage{amssymb}
\usepackage{mathtools}
\usepackage{amsthm}

\usepackage[capitalize,noabbrev]{cleveref}

\theoremstyle{plain}

\theoremstyle{definition}

\theoremstyle{remark}

\usepackage[textsize=tiny]{todonotes}

\icmltitlerunning{UGround: Towards Unified Visual Grounding with Unrolled Transformers}

\begin{document}

\twocolumn[
  \icmltitle{UGround: Towards Unified Visual Grounding with Unrolled Transformers}



  \icmlsetsymbol{equal}{*}

  \begin{icmlauthorlist}
    \icmlauthor{Rui Qian}{fudan}
    \icmlauthor{Xin Yin}{zju}
    \icmlauthor{Chuanhang Deng}{fudan,BEDI}
    \icmlauthor{Zhiyuan Peng}{sjtu} 
    \icmlauthor{Jian Xiong}{fudan}
    \icmlauthor{Wei Zhai}{fudan}
    \icmlauthor{Dejing Dou}{fudan,BEDI}
  \end{icmlauthorlist}

  \icmlaffiliation{fudan}{College of Computer Science and Artificial Intelligence, Fudan University}
  \icmlaffiliation{BEDI}{BEDI Cloud}
  \icmlaffiliation{zju}{Zhejiang University}
  \icmlaffiliation{sjtu}{Shanghai Jiao Tong University. \texttt{\{qiianruii,dengch2000\}@gmail.com}, \texttt{xyin@zju.edu.cn}, \texttt{pzy2000@sjtu.edu.cn}}

\icmlcorrespondingauthor{Dejing Dou}{dejingdou@gmail.com}

  \icmlkeywords{Large Multimodal Model, Visual Grounding}

  \vskip 0.2in
]



\printAffiliationsAndNotice{}  

\begin{abstract}
We present UGround, a \textbf{U}nified visual \textbf{Ground}ing paradigm that dynamically selects 
intermediate layers across \textbf{U}nrolled transformers as ``mask as prompt,'' diverging from the 
prevailing pipeline that leverages the fixed last hidden layer as ``\texttt{<SEG>} as prompt.'' UGround addresses two primary challenges posed by the prevailing paradigm: 
(1) its reliance on the fixed last hidden layer, which sequentially amplifies cumulative errors 
arising from layer-by-layer propagation without intermediate correction, and (2) its use of \texttt{<SEG>} as a prompt, 
which implicitly projects textual embeddings into visual space without explicit spatial cues 
(\eg, coordinates). Central to UGround is Policy-Prompted Masking, which comprises two key 
components: Stochastic Skip Connection (SSC) and Mask as Prompt (MasP). SSC is a reinforcement learning policy that, via stochastic sampling, 
allows each \texttt{<SEG>} token to slide across unrolled transformer layers, enabling dynamic layer selection at which it connects to the 
vision model (\eg, SAM) in a skip-connection fashion. Given the selected hidden layer, MasP uses the similarity map derived from the \texttt{<SEG>} token and image tokens as a soft logit 
mask to prompt SAM for mask generation, offering explicit spatial cues through its 
activation regions. To validate the effectiveness of UGround, we, for the first time, have unified 
visual grounding within a single framework from an attribute perspective, spanning from 
traditional refer expression segmentation to newly proposed reasoning segmentation, 
single-target to multi-target, positive query to false premise (empty target). 
All code and models are publicly available at {\hypersetup{urlcolor=colorlink}\href{https://github.com/rui-qian/UGround}{https://github.com/rui-qian/UGround}}.
\end{abstract}
\section{Introduction}
\label{sec:intro}
Visual grounding aims to align referring expressions with their corresponding 
regions in an image or video ~\citep{mao2016generation, yu2016modeling}. Despite 
existing efforts ~\citep{zou2023generalized, liu2023grounding, zou2023seem, liang2023open, 
liu2023gres, wang2022cris}, it remains largely unexplored under the 
newly emerging task settings ~\citep{lai2024lisa, wu2024see, xia2024gsva, rasheed2024glamm, 
ren2024pixellm, qian2024reasoning}. As shown in Table~\ref{tab:cmp_vis_grounding}, the shifts 
across recent tasks settings reveal how visual grounding has evolved, from referring to 
objects by an explicit mention (RES~\citep{rasheed2024glamm,ren2024pixellm}) to reasoning 
over descriptive language in an implicit fashion (RS~\citep{lai2024lisa,qian2026AnchorSeg});
from processing single-target queries per instance to handling multi-target scenarios 
(gRES~\citep{xia2024gsva}, Multi-RS~\citep{ren2024pixellm}); and from responding solely to positive queries to 
rejecting false premises with empty targets (FP-RES~\citep{wu2024see,xia2024gsva}).
Such shifts reflect the intrinsic properties of tasks, which we term ``attribute variation.''
\vspace{-0.4cm}
\begin{table}[htbp]
    \centering
    \caption{Attribute variation \wrt visual grounding. 
    RES: Referring Expression Segmentation, RS: Reasoning Segmentation, FP: False Premise, gRES: Generalized RES, Multi-RS: Multi-target RS.}
    \vspace{-0.1cm}\resizebox{1\linewidth}{!}
    {\tabcolsep=0.1cm
    \begin{tabular}{l|ccccc}
      \toprule
      Method & RES   & RS    & FP-RES & gRES  & Multi-RS \\
      \midrule
      LISA~\citep{lai2024lisa} & \Checkmark & \Checkmark & \textcolor[rgb]{ 1,  0,  0}{\ding{55}} & \textcolor[rgb]{ 1,  0,  0}{\ding{55}} & \textcolor[rgb]{ 1,  0,  0}{\ding{55}} \\
      SESAME~\citep{wu2024see} & \Checkmark & \Checkmark & \Checkmark & \textcolor[rgb]{ 1,  0,  0}{\ding{55}} & \textcolor[rgb]{ 1,  0,  0}{\ding{55}} \\
      READ~\citep{qian2024reasoning} & \Checkmark & \Checkmark & \Checkmark & \textcolor[rgb]{ 1,  0,  0}{\ding{55}} & \textcolor[rgb]{ 1,  0,  0}{\ding{55}} \\
      GLaMM~\citep{rasheed2024glamm} & \Checkmark & \Checkmark & \textcolor[rgb]{ 1,  0,  0}{\ding{55}} & \Checkmark & \textcolor[rgb]{ 1,  0,  0}{\ding{55}} \\
      GSVA~\citep{xia2024gsva}  & \Checkmark & \Checkmark & \Checkmark & \Checkmark & \textcolor[rgb]{ 1,  0,  0}{\ding{55}} \\
      PixelLM~\citep{ren2024pixellm} & \Checkmark & \Checkmark & \textcolor[rgb]{ 1,  0,  0}{\ding{55}} & \Checkmark & \Checkmark \\
      UGround (Ours) & \Checkmark & \Checkmark & \Checkmark & \Checkmark & \Checkmark \\
      \bottomrule
      \end{tabular}%
    }
\label{tab:cmp_vis_grounding}%
\vspace{-0.46cm}
\end{table}%

However, we observe that little research has looked into unifying existing tasks within a single 
framework from an attribute perspective, diverging from prior works~\citep{ren2024pixellm,rasheed2024glamm} 
that primarily focus on versatile capabilities. Generally, perception systems engaging 
with real-world scenarios are expected to reason over implicit instructions (cognition), interact with one or
multiple targets (generality), and appropriately reject queries when necessary (safety). Built upon  
attribute basis described above, versatility-level extensions, such as grounding in videos~\citep{wei2025hyperseg} or 
conversations~\citep{rasheed2024glamm}, find their footing. This leads us to 
ask: \textit{Can we design a unified architecture to bridge this gap?}
\begin{figure*}[h]
\begin{center}
\includegraphics[width=0.96\linewidth]{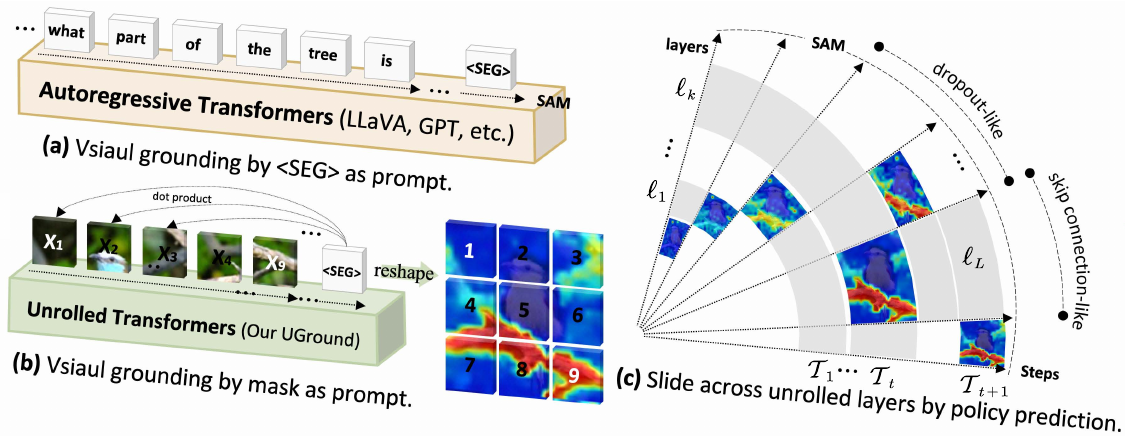}
\end{center}
\vspace{-0.4cm}
\caption{
Prior works typically use \texttt{<SEG>} token embeddings from the last hidden layer as the prompt in (a). 
In contrast, we leverage the similarity map in Eq.~(\ref{eq:similarity_map}), generated from \texttt{<SEG>} and image token embeddings across dynamically selected transformer layers, as the prompt in (b).  
Dynamic layer selection allows the similarity map to slide across transformer layers in (c).  Due to the sequential nature of transformers, $\ell_L-\ell_k$ layers are skipped, enabling a direct connection with SAM in a single forward step $\mathcal{T}_t$ (``skip-connection-like'').  
Across multiple forward steps $\mathcal{T}_1$ to $\mathcal{T}_T$, the connectivity to SAM varies 
, with only one path activated at each step (``dropout-like''). 
}
\label{fig:intro}
\vspace{-0.5cm}
\end{figure*}

Bearing this in mind, we revisit the methods of visual grounding presented in Table~\ref{tab:cmp_vis_grounding}, 
where we identify two key challenges of the off-the-shelf models: (1) its reliance on the fixed last hidden layer, 
and (2) its use of \texttt{<SEG>} as a prompt. Current Large Multimodal Models (LMMs) based visual grounding typically 
leverage \texttt{<SEG>} token stemming from the fixed last hidden layer of the vision-language model's stacked 
transformers (\eg, LLaVA) to prompt the downstream vision model (\eg, SAM) as shown in Fig.~\ref{fig:intro}(a). These stacked transformers can be deep, \eg, 
32 layers for LLaVA 7B, 40 layers for LLaVA 13B. In this paradigm, only the last hidden layer (layer 40) interacts with 
the vision SAM directly, whereas the other layers (layers 1-39) are separated. Consider telephone game by analogy, a message 
is passed from person to person (layer to layer) and becomes increasingly distorted before reaching the final 
participant (SAM). With only a single path, it sequentially amplifies cumulative errors arising from layer-by-layer 
propagation without any intermediate correction. In contrast to the similarity map with activation regions in Fig.~\ref{fig:intro}(b), 
\texttt{<SEG>} token, a text placeholder, lacks explicit spatial cues on its own, \eg, coordinates. It works via the alignment from text 
vocabulary into vision space in an implicit way of a fully connected layer. Inspired by the above findings, a natural idea 
is whether we can ``cheat'' this telephone game by letting the final participant (SAM) tap into intermediate 
layers (\eg, 1–39) in advance, through ``mask as prompt.''

To this end, we present UGround, a \textbf{U}nified visual \textbf{Ground}ing paradigm that dynamically selects 
intermediate layers across \textbf{U}nrolled transformers as ``mask as prompt.'' Central to our UGround is Policy-Prompted Masking (PPM), 
which consists of two key components: Stochastic Skip Connections (SSC) and Mask as Prompt (MasP). As shown in Fig.~\ref{fig:intro}(c), 
SSC is a stochastic reinforcement learning policy that allows each \texttt{<SEG>} token to slide across unrolled transformer layers, 
reinforcing dynamic layer selection at which it connects to the vision model (\eg, SAM) in a skip-connection fashion. 
Remarkably, SSC enables intermediate layers to skip subsequent ones and connect directly with SAM within each forward $\mathcal{T} _t$ 
(``skip-connection-like''). Across forward $\mathcal{T} _1$ to $\mathcal{T} _T$, such mechanism virtually connects all layers to SAM 
while activating only one pass at a time (``dropout-like''). Given the selected hidden layer, MasP uses the similarity map derived 
from the \texttt{<SEG>} token and image tokens as a soft logit mask to prompt SAM for mask generation, offering explicit spatial cues through its activation regions in the forward. Particularly, 
the similarity map is differentiable, beyond its implicit optimization via SAM backpropagation, we further impose a cross-entropy loss 
against the ground-truth mask to explicitly guide spatial cues in the backward. 
Our contributions are threefold:
\vspace{-0.3cm}
\begin{itemize}[leftmargin=0.3cm]
    \item We have ``unified'' visual grounding within a single framework from an attribute perspective. Previously, existing works 
    either focus on explicit rather than implicit expressions, or handle single-target instead of multi-target scenarios, or respond solely to positive 
    queries without rejecting false premises (empty target).
    \vspace{-0.2cm}
    \item We have ``unrolled'' stacked transformers, letting vision decoder tap into intermediate layers in a stochastic
    skip-connection fashion, in the way of ``mask as prompt.'' Importantly, we explicitly supervise the prompted mask (similarity map) against the ground-truth mask to further guide the model on where to attend.
    \vspace{-0.2cm}
    \item We conduct extensive experiments on the recently proposed ReasonSeg, the classic RefCOCO(+/g), and gRefCOCO(+/g) dataset. UGround outperforms the state of the art, achieving up to 9.0\% gains in cIoU on ReasonSeg and 12.1\% gains in N-acc on gRefCOCO(+/g) val sets.
\end{itemize}

\section{Related Work}
\textbf{Large Multimodal Models.}
Depending on the trade-offs between computational efficiency and performance ceiling, 
we categorize LMMs into two groups: (1) \textit{Fusion-centric models}, which aim for 
deeply fine-grained multimodal interaction by projecting visual, textual, and other modalities 
into a shared high-dimensional space with cross-attention mechanisms~\citep{alayrac2022flamingo,wang2023visionllm,detgpt,lv2023kosmos,zhang2023gpt4roi}.
Flamingo~\citep{alayrac2022flamingo} unlocks few-shot visual reasoning by bridging a frozen 
language model and a vision encoder with gated cross-attention. VisionLLM~\citep{wang2023visionllm} conditions a language model on image features 
via cross-attention to handle diverse vision-centric tasks, 
generating task-specific tokens for open-ended reasoning.
GPT4RoI~\citep{zhang2023gpt4roi} enables region-of-interest (RoI) reasoning through RoI-aware cross‑attention, allowing users to reference any image region in dialogue.
Kosmos-2~\citep{peng2024grounding} and DetGPT~\citep{detgpt} equip LMMs 
with grounding capabilities.
(2) \textit{Alignment-centric models}, which employ a divide-and-conquer strategy, 
projecting the outputs of frozen or lightly fine-tuned unimodal encoders (\eg, ViT for vision, 
LLaMA/GPT for text) into a shared latent space via a lightweight alignment adapter 
(\eg, projectors, Q-Formers)~\citep{li2023blip, ye2023mplug, li2025otter, liu2024visual, zhu2023minigpt}.
Otter~\citep{li2025otter} instruction-tunes a Flamingo-style vision–language model on the MIMIC-IT dataset.
LLaVA~\citep{liu2024visual} and MiniGPT-4~\citep{zhu2023minigpt} align a pretrained vision 
encoder with a frozen LLM using lightweight adapters for open-ended, instruction-following 
visual reasoning. Unlike LLaVA and MiniGPT-4, which rely on linear projections, BLIP-2~\citep{li2023blip} 
pioneers the Q-Former to distill visual features into compact queries 
for alignment with a frozen LLM. mPLUG-OWL~\citep{ye2023mplug} utilizes a visual 
abstractor to summarize global image features into condensed semantic tokens.
In contrast, our UGround builds upon LLaVA~\citep{liu2024visual} for visual autoregressive modeling and reasoning.

\textbf{Unified Segmentation Models.}
Depending on the semantic granularity of joint tasks, we divide literature into two groups: 
\textit{Versatility-oriented segmentation}, which aims to unify multiple functional 
properties within a single model~\citep{wei2025hyperseg, cheng2022masked, li2024omg, zhang2024omg, bai2024one, rasheed2024glamm}, 
as opposed to prior works dedicated exclusively to semantic~\citep{long2015fcn, ronneberger2015unet, 
badrinarayanan2017segnet}, instance~\citep{he2017mask, liu2018path, carion2020DETR}, 
or panoptic segmentation~\citep{kirillov2019panopticps, xiong2019upsnet, kirillov2019panopticfpn}. 
Mask2Former~\citep{cheng2022masked} pioneers a unified approach for image-level semantic, instance, and panoptic segmentation. OMG-Seg~\citep{li2024omg} extends Mask2Former to their video-level counterparts.  
HyperSeg~\citep{wei2025hyperseg} and OMG-LLaVA~\citep{zhang2024omg} distinguish themselves 
from previous purely visual models~\citep{cheng2022masked,li2024omg} by leveraging LMMs to generate interactive text.
VideoLISA~\citep{bai2024one} and HyperSeg~\citep{wei2025hyperseg} inject complex reasoning 
capabilities into generic video segmentation models. (2) \textit{Attribute-oriented segmentation}, 
which seeks to unify the intrinsic property variations within the task itself, spanning from explicit 
expressions to implicit instructions, one target to many targets, 
and positive queries to false premises. 
LISA~\citep{lai2024lisa} elevates RES to RS by leveraging self-reasoning capabilities of LMMs.
SESAME~\citep{wu2024see} and READ~\citep{qian2024reasoning} extend LISA by rejecting false premises.
GSVA~\citep{xia2024gsva} and PixelLM~\citep{ren2024pixellm} generalize RES to gRES by supporting 
multiple targets. Closest to our work, PixelLM~\citep{ren2024pixellm} cannot tackle 
empty targets, while GSVA~\citep{xia2024gsva} fails to handle multi-target reasoning.
Attribute-oriented unification is more fundamental, as it can be applied from a versatile 
perspective to any specialized task. 
Yet, we observe that little research has looked into unifying existing tasks within 
a single framework from an attribute perspective, we therefore present UGround.


\section{Reflection on Visual Grounding}
In this section, we first revisit visual grounding with LMMs and then analyze the underlying challenges posed by the typical paradigm presented in~\citep{lai2024lisa, wu2024see,qian2024reasoning}.
\subsection{Revisiting}
\begin{figure*}[t]
\begin{center}
\includegraphics[width=1\linewidth]{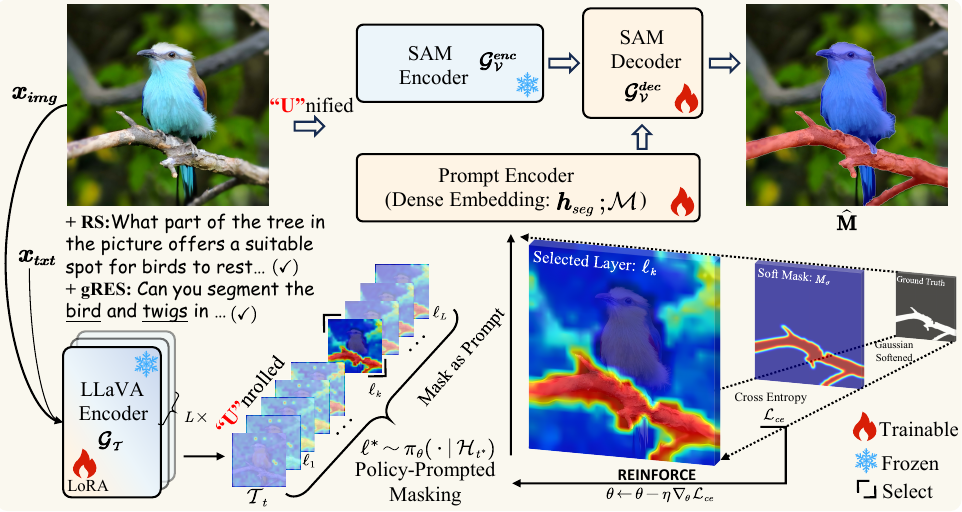}
\end{center}
\vspace{-0.3cm}
\caption{
Overview of our proposed UGround. Central to UGround is Policy-Prompted Masking (PPM), which stochastically selects layer $\ell^*$ among ``\textbf{U}nrolled'' transformers from  a policy distribution 
$\pi_\theta(\cdot \mid \mathcal{H}_{t^*})$ at which it connects to the 
vision model (\eg, SAM) in a skip-connection fashion. Given the layer $\ell^*$, MasP uses the similarity map as a soft logit mask to prompt SAM for mask generation, wherein we advance visual grounding 
within a ``\textbf{U}nified'' paradigm from an attribute perspective.
}
\label{fig:overview}
\vspace{-0.4cm}
\end{figure*}

\textbf{Problem Definition:}
Let $\mathbf{x}_{img} \in \mathbb{R}^{h \times w \times c}$ denote an input image with height $h$, 
width $w$, and $c$ channels, and let $\mathbf{x}_{txt}$ represent the paired textual input, 
which can span from an explicit mention (\eg, ``twigs") to an implicit description (\eg, 
``what part of the tree"). For segmentation, visual grounding aims to generate a mask 
that localizes the region of $\mathbf{x}_{img}$ described by the textual query $\mathbf{x}_{txt}$ as
\begin{align}
\begin{aligned}
    \Theta_{\mathrm{MLE}} = \arg\max_{\Theta} \; 
    \mathbb{E}_{\hat{\mathbf{M}} \sim \mathcal{G}_\theta(\cdot \mid \mathbf{x})} 
    \big[ \log \mathcal{G}_\theta(\hat{\mathbf{M}} \mid \mathbf{x}) \big],
\end{aligned}
\end{align}
where $\mathbf{x}$ denotes a multimodal sequence of discrete tokens derived from $\mathbf{x}_{img}$ and $\mathbf{x}_{txt}$.
$\hat{\mathbf{M}} \in \{0, 1\}^{h \times w}$, $1$ indicates the foreground and $0$ otherwise.
$\mathcal{G}_{\theta}$ indicates LMMs-based segmentation, including a multi-modal 
LLM $\mathcal{G} _{\mathcal{T}}$ and a visual backbone model $\mathcal{G} _{\mathcal{V}}$. Formally, 
$\mathcal{G} _{\theta}=\mathcal{G} _{\mathcal{V}} \circ \mathcal{G} _{\mathcal{T}}$, $\circ$ denotes the composition operation. 
As illustrated in Fig.~\ref{fig:overview}, we use LLaVA~\citep{liu2024visual} for $\mathcal{G} _{\mathcal{T}}$  and 
SAM~\citep{kirillov2023segment} for $\mathcal{G} _{\mathcal{V}}$, respectively. 

To equip $\mathcal{G}_{\mathcal{\theta}}$ with reasoning capabilities of LMMs, LISA~\citep{lai2024lisa} 
augments the text vocabulary of $\mathcal{G}_{\mathcal{T}}$ with a \texttt{<SEG>} token. 
Consider an input sequence $\mathbf{x}=({x}_{1},{x}_{2}, \dots, {x}_{T})$, 
$\mathcal{G} _{\mathcal{T}}$ consumes $\mathbf{x}$ as input with $L$ stacked transformer layers, which in turn response hidden states as 
\begin{align}
    \mathcal{H}^{(\ell)} &= \mathcal{G}_{\mathcal{T}}^{(\ell)}\big(\mathcal{H}^{(\ell-1)}\big), \quad \ell = 1, 2, \dots, L,
\end{align}
where $\mathcal{H}^{(0)}$ denotes the embeddings of the input sequence $\mathbf{x}$ at the embedding layer. 
The hidden state at the $\ell$-th layer can be presented as $\mathcal{H}^{(\ell )}=\left\{ \,\boldsymbol{h}_{1}^{(\ell )},\,\boldsymbol{h}_{2}^{(\ell )},\,\dots ,\,\boldsymbol{h}_{T}^{(\ell )}|\boldsymbol{h}_{t}^{(\ell )}\in \mathbb{R} ^d \right\}$.
Let $t^*$ be the position of the \texttt{<SEG>} token in the input sequence $\mathbf{x}$, 
then its corresponding hidden state at the $\ell$-th layer is $\boldsymbol{h}_{t^*}^{(\ell )}$, which is finally 
fed into $\mathcal{G} _{\mathcal{V}}^{dec}$ to generate the binary segmentation mask $\hat{\mathbf{M}}$ as
\begin{align}
    \begin{aligned}
        \hat{\mathbf{M}} = & \; \mathcal{G} _{\mathcal{V}}^{dec}(\mathbf{f}, \boldsymbol{h}_{seg}), \quad \boldsymbol{h}_{seg} = \varphi(\boldsymbol{h}^{(\ell )}_{t^*}) \in \mathbb{R} ^d,
        \label{eq:hatmask}
    \end{aligned}
\end{align}
where $\varphi(\cdot)$ is a multilayer perceptron (MLP) projection layer, $\mathbf{f}$ 
is the visual features of the input image $\mathbf{x}_{img}$ extracted by the vision backbone $\mathcal{G} _{\mathcal{V}}^{enc}$ in SAM~\citep{kirillov2023segment}.
However, we observe that existing works~\citep{lai2024lisa, wu2024see, xia2024gsva, rasheed2024glamm, 
ren2024pixellm, qian2024reasoning} predominantly utilize the embedding from the fixed last 
hidden layer ($\ell=L$) as the \texttt{<SEG>} token representation $\boldsymbol{h}_{seg}$, 
while the impact of intermediate layers remains largely unexplored.

\subsection{Analysis}
\label{sec:Analysis}
\textbf{Why Dynamic Layer Selection.} 
To analyze the impact of intermediate layers, we train UGround on the ReasonSeg \textit{train} 
set with two strategies: (1) relying solely on the last hidden layer ($\ell=L$), and (2) 
dynamically selecting from the intermediate layers ($\ell \in \{1,\dots,L-1\}$).
Qualitatively, we visualize the similarity maps across layers 1–40 of $\mathcal{G}_{\mathcal{T}}$ in Fig.~\ref{fig:layers1-40}. 
We observe that the fixed-last-layer strategy produces noisy similarity maps in intermediate layers (Fig.~\ref{fig:layers1-40}(a)), whereas the intermediate-layer counterpart yields a more discriminative foreground (Fig.~\ref{fig:layers1-40}(b)).
Quantitatively, we plot the predicted cIoU of the embeddings across layers 1–40. Fig.\textcolor{red}{2}(a) shows that every layer from 10–40 improves over the fixed-last-layer strategy by a clear and consistent margin. We further plot the per-layer similarity-map loss with respect to the ground-truth mask. Fig.\textcolor{red}{2}(b) shows that the intermediate-layer strategy begins converging from layer 19 and ultimately reaches a lower loss level, while the fixed-last-layer strategy only starts converging from layer 28, indicating that dynamic layer selection accelerates convergence.
Dynamic layer selection provides insight into the behavior of intermediate layers. By allowing these layers to directly participate in interactions and joint training with SAM, akin to ``cheating" in a phone game (\ie, receiving intermediate hints), each layer's performance is improved, ultimately resulting in an overall enhancement.
\begin{figure}[htbp]
    \centering    \includegraphics[width=1.01\linewidth]{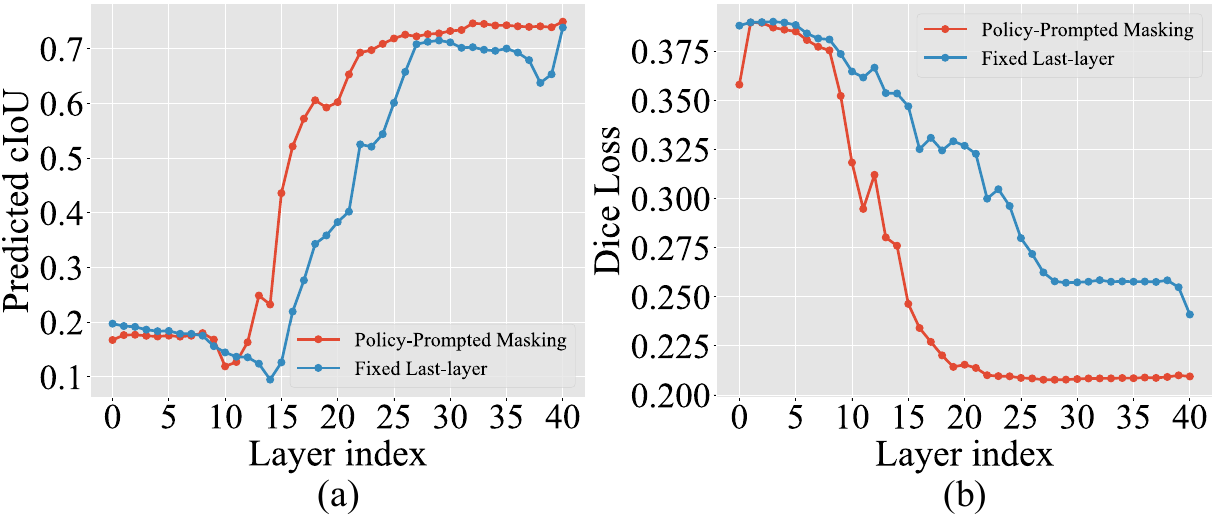}
    \label{fig:ppm}
    \vspace{-0.4cm}
    \caption{(a) shows that the dynamic layer selection strategy outperforms the fixed last hidden layer strategy across all middle layers. (b) plots the loss of similarity maps against the soft ground-truth mask, demonstrating faster convergence at intermediate layers.
}
\vspace{-0.3cm}
\end{figure}%


\textbf{Why Similarity as Mask.} 
To investigate whether the similarity map can serve as a prompt, we conduct two analyses. First, to measure the consistency ($\mathcal{M}_{\mathrm{IoU}}$) between the similarity map and the ground-truth mask, we compute the grid search-based Intersection over Union (IoU) following READ~\citep{qian2024reasoning}. Second, to probe whether SAM can understand the semantics encoded in the similarity map, we directly use the similarity map as a logit mask to prompt the original SAM~\citep{kirillov2023segment} (without any additional training), denoted by $\mathcal{M}_{\mathrm{prompt}}$. Table~\ref{tab:Analysis} shows that SAM can potentially leverage the similarity map as a prompt ($17.0\%$ vs. $48.4\%$). Further, when the similarity map is directly converted into a binary mask for prediction, its performance measured by $\mathcal{M}_{\mathrm{IoU}}$ can even rival that of the trained \texttt{<SEG>}\textsubscript{prompt}, as evidenced by a 4.3\% higher cIoU on the ReasonSeg \textit{test} set ($35.0\% \ \text{vs.}\ 30.7\%$) in the $3^{\text{rd}}$ row.

\begin{table}[htbp]
  \centering
  \caption{
  Quantitative analysis of the similarity maps on the ReasonSeg \textit{test} split. 
  $\mathcal{M}$\textsubscript{prompt} indicates mask as a prompt for the original SAM, 
while $\mathcal{M}$\textsubscript{IoU} quantifies the overlap between the similarity map and the ground-truth mask. 
\texttt{<SEG>}\textsubscript{prompt} refers to \texttt{<SEG>} token as prompt for SAM.
}
\resizebox{\linewidth}{!}
{\tabcolsep=0.1cm
    \begin{tabular}{l|cc|cc|cc}
    \toprule
    \multirow{2}[4]{*}{Method} & \multicolumn{2}{c|}{$\mathcal{M}$\textsubscript{prompt}} & \multicolumn{2}{c|}{$\mathcal{M}$\textsubscript{IoU}}& \multicolumn{2}{c}{\texttt{<SEG>}\textsubscript{prompt}} \\
\cmidrule{2-7}          & gIoU  & cIoU  & gIoU  & cIoU & gIoU  & cIoU \\
    \midrule
    LISA-7B~\citep{lai2024lisa}  & {16.1}  & {17.0}  & 32.2  & 35.6 & {47.3} & {48.4} \\
    GSVA-7B~\citep{xia2024gsva} & 15.0  & 14.7    & {33.2}  & {37.4} & {42.8} & {43.8}\\
    SESAME-7B~\citep{wu2024see} & 15.6  & 15.6    & 31.6  & 35.0 & 34.9 & 30.7\\    
    \bottomrule
    \end{tabular}%
}
 \label{tab:Analysis}%
 \vspace{-0.3cm}
\end{table}%

\noindent \textbf{Summary}. Analysis of the similarity maps reveals the following: (a) 
Similarity maps in LMMs are highly consistent with the ground-truth mask, suggesting that 
they can potentially serve as prompts for generating the segmentation mask.
(b) Dynamic layer selection allows the intermediate layers to directly interact with SAM, 
resulting in a more discriminative representation, as reflected by the activated regions 
in the similarity maps. 
Motivated by the insights from the similarity maps, we leverage it in three ways: as a logit 
mask to prompt SAM (prompt), as a loss constraint to guide the model on where to ``attend'' 
(constraint), and as a policy reward for dynamic layer selection (signal) 
(see Appendix~\ref{app:layers1-40} for further analysis).
\section{Proposed UGround}
In this section, we present UGround, which dynamically selects intermediate layers across unrolled 
transformers and connects them to SAM in a skip-connection fashion, where the similarity map serves 
as a prompt, thereby contributing to a unified visual grounding paradigm. As depicted in Fig.~\ref{fig:overview}, 
central to UGround is Policy-Prompted Masking (PPM), consisting of two components: Stochastic Skip 
Connection (SSC) and Mask as Prompt (MasP). SSC is a reinforcement learning policy that stochastically 
allows each \texttt{<SEG>} token to slide across unrolled transformer layers, dynamically selecting 
where it connects to the vision model (\eg, SAM) via a skip connection. MasP, given the selected layer, 
leverages the similarity map between the \texttt{<SEG>} token and image tokens as a soft logit mask 
to prompt SAM, providing explicit spatial cues through its activation regions.
As the key novelty of our work is PPM, we present it first in Sec.~\ref{method:arch}, then 
the training objectives in Sec.~\ref{TrainingObjectives}.

\subsection{Policy-Prompted Masking}
\label{method:arch}

\textbf{Stochastic Skip Connection.} 
Prior works~\citep{lai2024lisa, wu2024see, xia2024gsva, rasheed2024glamm, 
ren2024pixellm, qian2024reasoning} empirically use the embedding from the fixed last hidden 
layer ($\ell=L$) as the \texttt{<SEG>} token representation $\boldsymbol{h}_{seg}$.
In contrast, we exploit intermediate layers ($\ell \in \{1,\dots,L-1\}$) to derive more 
discriminative semantics in a skip-connection fashion. To this end, we formulate the problem 
of selecting a layer for the \texttt{<SEG>} token as a reinforcement learning task. 
Each layer selection is treated as an action $a \equiv \ell^*$ taken in the state $s \equiv \boldsymbol{h}_{t^*}$, 
which corresponds to the unrolled \texttt{<SEG>} embeddings across layers. The agent receives a reward $r$ that 
reflects the consistency of the similarity map with the ground-truth mask. Specifically, let 
$\mathcal{H}_{t^*}$ 
represent the hidden states of the \texttt{<SEG>} token from all layers ($\ell \in \{1,\dots,L\}$) 
at position $t^*$ of the input sequence $\mathbf{x}$ as 
$\mathcal{H}_{t^*} = \{ \boldsymbol{h}_{t^*}^{(1)}, \boldsymbol{h}_{t^*}^{(2)}, \dots, \boldsymbol{h}_{t^*}^{(L)} \} \in \mathbb{R}^{L \times d}$, 
let $\mathbf{w} \in \mathbb{R}^{L \times d}$ denote learnable weights. The policy distribution 
$\pi_\theta(\cdot \mid \mathcal{H}_{t^*})$ is defined as
\begin{align}
\pi_\theta(\ell \mid \mathcal{H}_{t^*}) = \frac{\exp(s_\ell)}{\sum_{j=1}^{L} \exp(s_j)}, s_\ell = \boldsymbol{h}_{t^*}^{(\ell)} \cdot \mathbf{w}_\ell,
\vspace{-0.3cm}
\end{align}
where $s_\ell$ is the logit scores for each layer. $\boldsymbol{h}_{t^*}^{(\ell)}$ is the \texttt{<SEG>} embedding at layer $\ell$ 
and $\mathbf{w}_\ell$ is the corresponding learnable weight. The agent chooses a layer $\ell$ 
according to a categorical distribution parameterized by learnable weights $\mathbf{w}$.
During training, the layer $\ell^*$ is sampled from this distribution to allow exploration as $\ell^* \sim \pi_\theta(\cdot \mid \mathcal{H}_{t^*}),$
To reduce variance, we introduce a baseline 
$b$, which can be an EMA-smoothed reward or a critic network prediction. The actual reward $r$ 
is derived from the similarity map, as described in 
Algorithm~\hyperref[alg:ppm]{(\textcolor{red}{\ref*{alg:ppm}})}
steps 5-8. We 
therefore formulate the REINFORCE gradient as
\vspace{-0.1cm}
\begin{align}
\!\nabla_\theta J(\theta)\!=\!\mathbb{E}_{\ell^* \sim \pi_\theta(\cdot \mid \mathcal{H}_{t^*})} 
\Big[ (r\!-\!b) \, \nabla_\theta \log \pi_\theta(\ell^*\!\mid \mathcal{H}_{t^*}) \Big],
\end{align}
where the log-probability of the sampled action $a \equiv \ell^*$ is weighted by the advantage $r - b$. To get the reward $r$, we first compute the similarity map $\mathcal{M}$ between 
the \texttt{<SEG>} token and image tokens at layer $\ell$ and its corresponding soft mask $\mathcal{M}_\sigma$. 
Consider a sequence of discrete image tokens $\mathbf{z} = (z_1, z_2, \dots, z_k)$ in the input sequence $\mathbf{x}$, 
where $k$ is the number of image tokens. The image token embeddings across layers are then denoted as
\vspace{-0.1cm}
$\mathcal{H}_{\mathbf{z}} = \{ \boldsymbol{h}_z^{(\ell)} \mid z \in \mathbf{z}, \ \ell = 1, \dots, L \} \in \mathbb{R}^{k \times L \times d}$,
where \(\boldsymbol{h}_z^{(\ell)}\) is the embedding of the $z$-th image token at layer \(\ell\). 
Given the \texttt{<SEG>} token embedding $\boldsymbol{h}_{t^*}^{(\ell)} \in \mathbb{R}^d$ 
and the image token embeddings $\mathcal{H}_{\mathbf{z}}^{(\ell)} \in \mathbb{R}^{k \times d}$ at the 
selected layer \(\ell\), the similarity scores can be computed as 
\begin{align}
    \mathcal{S}^{(\ell)}\!=\!\mathcal{H}_{\mathbf{z}}^{(\ell)} \, {h_{t^*}^{(\ell)}}^\top \in \mathbb{R}^{k},
    \label{eq:similarity_map}
\end{align}
where each element $\mathcal{S}_i^{(\ell)} = (h_{z_i}^{(\ell)})^\top h_{t^*}^{(\ell)}$ represents 
the similarity between the \texttt{<SEG>} token and the \(i\)-th selected image token at layer \(\ell\).
$\mathcal{S}^{(\ell)}$ are arranged in a 2D grid, and interpolated to obtain the soft logit mask $\mathcal{M} \in [0,1]^{H \times W}$.
Let the ground-truth mask be $M \in \{0,1\}^{H \times W}$, which we regard as a function
$M : \{1,\dots,H\} \times \{1,\dots,W\} \to \{0,1\}$ over the discrete pixel grid.
We extend $M$ to the continuous domain $\mathbb{R}^2$ by assigning each pixel a constant
value on its corresponding region in the plane.
Applying Gaussian smoothing to this extended function yields a continuous-valued function
$M_\sigma : \mathbb{R}^2 \to [0,1]$, where $M_\sigma$ serves as the soft label (heatmap). Formally, $M_\sigma$ can be expressed as a convolution integral as $M_\sigma(u,v) = \iint_{\mathbb{R}^2} M(x,y) \, G_\sigma(u-x, v-y) \, dx \, dy,$  
where $G_\sigma(x, y) \propto \exp(- \frac{x^2 + y^2}{2\sigma^2})$, is a 2D Gaussian kernel with standard deviation $\sigma$. 
The reward for REINFORCE is computed using the similarity map $\mathcal{M}$ and the soft label $M_\sigma$ as $r = - \big( \mathcal{L}_{bce}(\mathcal{M}, M_\sigma) + \mathcal{L}_{dice}(\mathcal{M}, M_\sigma) \big),$
where \(\mathcal{L}_{bce}\) and \(\mathcal{L}_{dice}\) denote the Gaussian BCE and Dice losses defined over the similarity map $\mathcal{M}$.
The policy loss is then defined as
\begin{align}
\mathcal{L}_{\mathrm{policy}} = - (r - b_t) \, \log \pi_{\theta}(\ell^* \mid \mathcal{H}_{t^*}).
\end{align}
Algorithm~\hyperref[alg:ppm]{(\textcolor{red}{\ref*{alg:ppm}})}
outlines a single forward pass of the PPM. The transformer layers are stacked sequentially. By skipping the subsequent \( L - \ell^* \) layers, the PPM allows layer \( \ell^* \) to connect directly to SAM in a skip-connection manner, thereby mitigating cumulative errors that arise from layer-by-layer propagation through intermediate correction (\ie, a shortcut that bypasses subsequent layers). Across multiple forward passes (from \(1\) to \(\mathcal{T}\)), the PPM dynamically alters the model's internal connectivity, analogous to the dropout mechanism: although all layers are virtually connected to SAM, only one path is activated per pass. This stochastic sampling can be interpreted as a form of Monte Carlo uncertainty estimation~\citep{gal2016dropout}, which quantifies predictive uncertainty via multiple stochastic forward passes. In the context of PPM, each forward pass activates a different layer-path to SAM, and the ensemble 
\begin{algorithm}[htbp]
    \caption{Policy-Prompted Masking Algorithm}
    \label{alg:ppm}
    \small
    \begin{algorithmic}[1]
        \REQUIRE 
        Given unrolled transformer layers with $L$ layers, $\mathcal{H}_{t^*} \in \mathbb{R}^{L \times d}$ is hidden states 
        of \texttt{<SEG>} token, $\mathcal{H}_{\mathbf{z}} \in \mathbb{R}^{k \times L \times d}$ is hidden states of image tokens; 
        $\mathbf{w}$ is learnable weights $\mathbf{w} \in \mathbb{R}^{L \times d}$;$M_\sigma : \mathbb{R}^2 \to [0,1]$, where $M_\sigma$ is 
        the soft heatmap, with a Gaussian kernel $G_\sigma(x,y)$;
        $r$ is the reward score, $b$ is the baseline, $M \in \{0,1\}^{H \times W}$ is ground-truth mask;$\mathcal{T}$ denotes the number of forward passes, each of which corresponds 
        to an independent policy sampling followed by a mask prediction.
        \ENSURE 
        Selected layer $\ell^*_t$ and similarity map $\mathcal{M}_t$
        \FOR{$t = 1, \dots, \mathcal{T}$} 
        \STATE $s_\ell = \boldsymbol{h}_{t^*}^{(\ell)} \cdot \mathbf{w}_\ell$ for $\ell = 1, \dots, L$
        \STATE $\pi_\theta(\ell \mid \mathcal{H}_{t^*}) = \frac{\exp(s_\ell)}{\sum_{j=1}^{L} \exp(s_j)}$
        \STATE $\ell^* \sim \pi_\theta(\cdot \mid \mathcal{H}_{t^*})$
        \STATE $\mathcal{S}^{(\ell^*)} = \mathcal{H}_{\mathbf{z}}^{(\ell^*)} \, {h_{t^*}^{(\ell^*)}}^\top \in \mathbb{R}^{k}$
        \STATE $\mathcal{M} \in [0,1]^{H \times W} \;\leftarrow\; \mathcal{S}^{(\ell^*)}$
        \STATE $M_\sigma(u,v) = \iint_{\mathbb{R}^2} M(x,y) \, G_\sigma(u-x, v-y) \, dx \, dy$
        \STATE $r = - \big( \mathcal{L}_{bce}(\mathcal{M}, M_\sigma) + \mathcal{L}_{dice}(\mathcal{M}, M_\sigma) \big)$
        \STATE $b_t = \alpha b_{t-1} + (1-\alpha) r$
        \STATE $\mathcal{L}_{\mathrm{policy}} = -(r - b_t) \log \pi_\theta(\ell^* \mid \mathcal{H}_{t^*})$
        \STATE $\theta \leftarrow \theta - \eta \, \nabla_\theta \mathcal{L}_{\mathrm{policy}}$
        \ENDFOR
        \RETURN $\{\ell^*_t, \mathcal{M}_t\}_{t=1}^{\mathcal{T}}$
    \end{algorithmic}
\end{algorithm}

of these paths (\ie, 32 in LLaVA) alleviates over-reliance on any single trajectory,  enhancing the model's robustness.

\textbf{Mask as Prompt.}
Compared with \texttt{<SEG>} tokens, the similarity map provides more explicit spatial cues, 
as analyzed in Sec.~\ref{sec:Analysis}. We use the similarity map as a logit mask to prompt SAM. 
Accordingly, we reformulate Eq.~(\ref{eq:hatmask}) as 
$\hat{\mathbf{M}}=\;\mathcal{G} _{\mathcal{V}}^{dec}(\mathbf{f},  \boldsymbol{h}_{seg}, \mathcal{M} ).$
The similarity map is continuously differentiable, allowing its gradients to be backpropagated 
through the SAM module for implicit optimization. To further guide the model to focus on target 
regions, we supervise the similarity map $\mathcal{M}$ by imposing a constraint (\eg, using binary cross-entropy (BCE) or Dice loss) 
against a soft ground-truth mask $M_\sigma$, providing explicit guidance on where the model should ``attend.''

\subsection{Training Objectives}
\label{TrainingObjectives}
Apart from the text generation loss $\mathcal{L}_{txt}$ in $\mathcal{G}_{\mathcal{T}}$, and the 
segmentation mask loss $\mathcal{L}_{mask}$ in $\mathcal{G}_{\mathcal{V}}^{dec}$~\citep{lai2024lisa,wu2024see},
we use policy loss $\mathcal{L}_{\mathrm{policy}}$ for stochastic skip connection, 
and binary 
cross-entropy (BCE) loss and Dice loss for similarity map as 
\begin{align}
\begin{aligned}
    \mathcal{L}_{\mathrm{policy}}&= - (r - b_t) \, \log \pi_{\theta}(\ell^* \mid \mathcal{H}_{t^*}),
\\
\mathcal{L} _{\mathcal{M}}=\lambda _{bce}\mathcal{L} _{bce}&(\mathcal{M}, M_\sigma)+\lambda _{dice}\mathcal{L} _{dice}(\mathcal{M}, M_\sigma),
\label{eq:losses}
\end{aligned}
\end{align}
where $M_\sigma$ is the soft ground-truth target, $\lambda$ denotes the loss weight, 
$\mathcal{M}$ is the similarity map. The overall objective $\mathcal{L}$ summarizes those losses 
in Eq.~(\ref{eq:losses}), weighted by $\lambda_{txt}$, $\lambda_{mask}$,  $\lambda_{\mathcal{M}}$, and $\lambda_{policy}$ as
\begin{equation}
   \!\!\!\mathcal{L}\! =\! \lambda_{txt} \mathcal{L}_{txt}\! +\! \lambda_{mask} \mathcal{L}_{mask}\! +\! \lambda_{\mathcal{M}} \mathcal{L}_{\mathcal{M}}\! + \!\lambda_{policy} \mathcal{L}_{\mathrm{policy}}.
\end{equation}

\begin{table*}[htbp]
    \centering
    \caption{
    Benchmarking reasoning segmentation models on the ReasonSeg dataset, sorted in ascending order of cIoU on the test set. *: reproduced from official models. ft: fine-tuned on 239 samples.
}
\vspace{-0.1cm}
    \resizebox{0.97\linewidth}{!}
    {
    \label{table:reason_seg}   
    \tabcolsep=0.5cm
    {
        \begin{tabular}{l|cc|cc|cc|cc}
            \toprule
            
            \multirow{3}*{Method} & \multicolumn{2}{c|}{val} & \multicolumn{6}{c}{test} \\ 
            
            \specialrule{0em}{0pt}{1pt}
            \cline{2-9}
            \specialrule{0em}{0pt}{1pt}
            
            
            ~ & \multicolumn{2}{c|}{overall} & \multicolumn{2}{c|}{short query} & \multicolumn{2}{c|}{long query} & \multicolumn{2}{c}{overall} \\

            \specialrule{0em}{0pt}{1pt}
            \cline{2-9}
            \specialrule{0em}{0pt}{1pt}
            
            ~ & gIoU & cIoU & gIoU & cIoU & gIoU & cIoU & gIoU & cIoU \\ 
            
            \specialrule{0em}{0pt}{1pt}
            \hline
            \specialrule{0em}{0pt}{1pt}
            X-Decoder~\citep{zou2023generalized} & 22.6 & 17.9 & 20.4 & 11.6 & 22.2 & 17.5 & 21.7 & 16.3 \\

            Grounded-SAM~\citep{liu2023grounding} & 26.0 & 14.5 & 17.8 & 10.8 & 22.4 & 18.6 & 21.3 & 16.4 \\

            SEEM~\citep{zou2023seem} & 25.5 & 21.2 & 20.1 & 11.5 & 25.6 & 20.8 & 24.3 & 18.7 \\
            
            OVSeg~\citep{liang2023open} & 28.5 & 18.6 & 18.0 & 15.5 & 28.7 & 22.5 & 26.1 & 20.8  \\

            GRES~\citep{liu2023gres} & 22.4 & 19.9 & 17.6 & 15.0 & 22.6 & 23.8 & 21.3 & 22.0 \\    %
    
            \specialrule{0em}{0pt}{1pt}
            \hline
            \specialrule{0em}{0pt}{1pt}
            
            *SESAME~\citep{wu2024see} & {40.3} & {41.6} & {28.9} & {26.3} & {37.3} & {31.9}& {34.9} & {30.7} \\

            LLaVA1.5-7B + OVSeg~\citep{lai2024lisa} & 38.2 & 23.5 & 24.2 & 18.7 & 44.6 & 37.1 & 39.7 & 31.8 \\

            *GSVA~\citep{xia2024gsva} & {45.6} & {41.5} & {37.9} & {36.5} & {44.3} & {46.0}& {42.8} & {43.8} \\

            *PixelLM~\citep{ren2024pixellm} & {49.7} & {49.6} & {39.5} & {38.8} & {49.5} & {45.6}& {47.1} & {44.3} \\

            LISA-7B~\citep{lai2024lisa} & 52.9 & 54.0 & 40.6 & 40.6 & 49.4 & 51.0 & 47.3 & 48.4 \\

            HyperSeg-3B~\citep{wei2025hyperseg} & 59.2 & 56.7 & - & - & - & - & - & - \\

            VISA-7B~\citep{yan2024visa} & 52.7 & 57.8 & - & - & - & - & - & - \\

            VideoLISA-3.8B~\citep{bai2024one} & 61.4 & 67.1 & 43.8 & 42.7 & 56.9 & 57.7 & 53.8 & 54.4 \\

            LISA-7B-LLaVA1.5 (ft)~\citep{lai2024lisa} & 61.3 & 62.9 & {48.3} & {46.3} & 57.9 & 59.7 & 55.6 & 56.9 \\
            
            READ-7B-LLaVA1.5 (ft)~\citep{qian2024reasoning} & 59.8 & 67.6 & 52.6 & 49.5  & 60.4 & 61.0 & 58.5 & 58.6 \\
            
            LISA++-7B-LLaVA1.5 (ft)~\citep{yang2023lisa++} & 64.2 & 68.1 & {49.6} & \textbf{51.1} & 59.3 & 61.7 & 57.0 & 59.5 \\

            RSVP-GPT~\citep{lu2025rsvp} & 64.7 & 63.1 & \textbf{55.4} & 50.4 & 61.9 & 62.5 & 60.3 & 60.0 \\
            \rowcolor{mygray}
            UGround-7B-LLaVA1.5 (ft) & \textbf{66.1} & \textbf{72.1} & 55.1 & 48.5  & \textbf{66.3} & \textbf{70.2} & \textbf{63.6} & \textbf{65.4} \\
            \specialrule{0em}{0pt}{1pt}
            \hline
            \specialrule{0em}{0pt}{1pt}
            LLaVA1.5-13B + OVSeg~\citep{lai2024lisa} & 37.9 & 26.4 & 27.1 & 19.4 & 46.1 & 40.6 & 41.5 & 34.1 \\
            
            LISA-13B-LLaVA1.5~\citep{lai2024lisa} & 57.7 & 60.3 & 50.8 & 50.0 & 54.7 & 50.9 & 53.8 & 50.8 \\
               
            LISA-13B-LLaVA1.5(ft)~\citep{lai2024lisa} & 65.0 & 72.9 & 55.4 & 50.6 & 63.2 & 65.3 & 61.3 & 62.2 \\
            READ-13B-LLaVA1.5 (ft)~\citep{qian2024reasoning} & - & - & 55.4 & \textbf{53.7}  & 64.4 & 65.1 & 62.2 & 62.8 \\
            \rowcolor{mygray}
            UGround-13B-LLaVA1.5 (ft) & \textbf{67.9} & \textbf{74.9} & \textbf{57.2} & 50.9 & \textbf{67.5} & \textbf{69.4} & \textbf{65.0} & \textbf{65.5} \\
            \bottomrule            
        \end{tabular}
    }  
    }  
        \vspace{-0.13cm}
\end{table*}
\begin{table*}[htbp]
    \centering    
    \caption{
    Benchmarking referring segmentation models on the RefCOCO(+/g) dataset,  sorted in ascending order of cIoU on the RefCOCOg val set. 
    }
    \vspace{-0.2cm}
    \resizebox{\linewidth}{!}
    {
    \label{table:refer_seg}   
    \tabcolsep=0.7cm
    {
        \begin{tabular}{l|ccc|ccc|cc}
            \toprule
            
            \multirow{2}*{Method} & \multicolumn{3}{c|}{RefCOCO} & \multicolumn{3}{c|}{RefCOCO+}  & \multicolumn{2}{c}{RefCOCOg} \\ 
            
            \specialrule{0em}{0pt}{1pt}
            \cline{2-9}
            \specialrule{0em}{0pt}{1pt}
            
            ~ & val & testA & testB & val & testA & testB & val(U) & test(U) \\ 
                 
            \specialrule{0em}{0pt}{1pt}
            \hline
            \specialrule{0em}{0pt}{1pt}

            MCN~\citep{luo2020multi} & 62.4 & 64.2 & 59.7 & 50.6 & 55.0 & 44.7 & 49.2 & 49.4 \\

            VLT~\citep{ding2021vision} & 67.5 & 70.5 & 65.2 & 56.3 & 61.0 & 50.1 & 55.0 & 57.7 \\

            CRIS~\citep{wang2022cris} & 70.5 & 73.2 & 66.1 & 62.3 & 68.1 & 53.7 & 59.9 & 60.4 \\

            LAVT~\citep{yang2022lavt} & 72.7 & 75.8 & 68.8 & 62.1 & 68.4 & 55.1 & 61.2 & 62.1 \\
                        
            X-Decoder~\citep{zou2023generalized} & - & - & - & - & - & - & 64.6 & -  \\

            ReLA~\citep{liu2023gres} & 73.8 & 76.5 & 70.2 & {66.0} & {71.0} & 57.7 & 65.0 & 66.0 \\

            SEEM~\citep{zou2023seem} & - & - & - & - & - & - & 65.7 & -  \\


            \midrule

            Segment Anyword~\citep{liu2025segment} & {55.3} & {47.9} & {66.0} & 55.6 & {47.4} & {67.0} & {58.4} & {60.1} \\

            VISA-7B~\citep{yan2024visa} & {72.4} & {75.5} & {68.1} & 59.8 & 64.8 & 53.1 & 65.5 & 66.4 \\
            
            SESAME~\citep{wu2024see} & {74.7} & {-} & {-} & 64.9 & {-} & {-} & {66.1} & {-} \\

            LISA-7B~\citep{lai2024lisa} & {74.9} & {79.1} & {72.3} & 65.1 & 70.8 & {58.1} & {67.9} & {70.6} \\

            CoRes~\citep{bao2024cores} & {76.0} & {78.6} & {72.5} & {65.1} & {70.0} & {58.6} & {69.0} & {70.7} \\

            PixelLM-7B~\citep{ren2024pixellm} & {73.0} & {76.5} & {68.2} & 66.3 & 71.7 & {58.3} & {69.3} & {70.5} \\

            READ-7B~\citep{qian2024reasoning} & 78.1 &  80.2 & 73.2 & 68.4 & 73.7 & 60.4 & 70.1 & 71.4 \\

            SegLLM-7B~\citep{wang2024segllm} & {80.2} & {81.5} & {75.4} & 70.3 & 73.0 & {62.5} & {72.6} & {73.6} \\
            
            GSVA-7B~\citep{xia2024gsva} & {77.2} & {78.9} & {73.5} & 65.9 & 69.6 & {59.8} & {72.7} & {73.3} \\
            
            OMG-LLaVA~\citep{zhang2024omg} & {78.0} & {80.3} & {74.1} & 69.1 & 73.1 & 63.0 & 72.9 & 72.9 \\
            
            GLaMM-7B~\citep{rasheed2024glamm} & {79.5} & {83.2} & {76.9} & 72.6 & \textbf{78.7} & {64.6} & {74.2} & {74.9} \\
            
            
            
            \rowcolor{mygray}
            UGround-7B & \textbf{80.6} &  \textbf{83.5}& \textbf{77.7} & \textbf{72.8} & 77.5 &\textbf{65.6} & \textbf{74.7} & \textbf{76.1} \\
            
            \bottomrule            
        \end{tabular}
    }
    }
    \vspace{-0.2cm}
\end{table*}
\begin{table*}[htbp]
    \centering
    \caption{
    Benchmarking generalized referring expression segmentation (GRES) models on the gRefCOCO~\citep{liu2023gres} dataset, sorted in ascending order of cIoU on the gRefCOCO validation set. Values are taken from~\citep{liu2023gres}.  N-acc.: the accuracy of correct null-target classification. ft: fine-tuned on gRefCOCO training set.
    }
    \vspace{-0.1cm}
    \tabcolsep=0.6cm
    \resizebox{\linewidth}{!}{
    \begin{tabular}{l|ccc|ccc|ccc}
    \toprule
    \multirow{2}{*}{Method} & \multicolumn{3}{c|}{Validation Set} & \multicolumn{3}{c|}{Test Set A}  & \multicolumn{3}{c}{Test Set B} \\
    \specialrule{0em}{0pt}{1pt}
    \cline{2-10}
    \specialrule{0em}{0pt}{1pt}
    & gIoU & cIoU & N-acc. & gIoU & cIoU & N-acc. & gIoU & cIoU & N-acc. \\
    \midrule
    MattNet~\citep{yu2018mattnet} & 48.24 & 47.51 & 41.15 & 59.30 & 58.66 & 44.04 & 46.14 & 45.33 & 41.32 \\
    LTS~\citep{jing2021locate} & 52.70 & 52.30 & - & 62.64 & 61.87 & - & 50.42 & 49.96 & - \\
    VLT~\citep{ding2021vision} & 52.00 & 52.51 & 47.17 & 63.20 & 62.19 & 48.74 & 50.88 & 50.52 & 47.82 \\
    CRIS~\citep{wang2022cris} & 56.27 & 55.34 & - & 63.42 & 63.82 & - & 51.79 & 51.04  & - \\
    LAVT~\citep{yang2022lavt} & 58.40 & 57.64 & 49.32 & 65.90 & 65.32 & 49.25 & 55.83 & 55.04 & 48.46 \\
    ReLA~\citep{liu2023gres} & 63.60 & 62.42 & 56.37 & 70.03 & 69.26 & 59.02 & 61.02 & 59.88 & 58.40 \\
    \midrule
    LISA-Vicuna-7B~\citep{lai2024lisa} & 32.21 & 38.72 & 2.71 & 48.54 & 52.55 & 6.37 & 39.65 & 44.79 & 5.00 \\
    GSVA-Vicuna-7B~\citep{xia2024gsva} & 63.32 & 61.70 & 56.45 & 70.11 & 69.23 & 63.50 & 61.34 & 60.26 & 58.42  \\
    LISA-Vicuna-7B (ft)~\citep{lai2024lisa} & 61.63 & 61.76 & 54.67 & 66.27 & 68.50 & 50.01 & 58.84 & 60.63 & 51.91 \\
    GSVA-Vicuna-7B (ft)~\citep{xia2024gsva} & 66.47 & 63.29 & 62.43 & 71.08 & 69.93 & 65.31 & 62.23 & 60.47 & 60.56 \\
    \rowcolor{mygray}
    UGround-LLaVA1.5-7B(ft) & \textbf{72.46} & \textbf{65.56} & \textbf{74.53} & \textbf{74.29} & \textbf{70.87} & \textbf{73.93} & \textbf{66.85} & \textbf{61.84} & \textbf{71.22} \\
    \bottomrule
    \end{tabular}}
    \label{tab:gres}
\end{table*}
\clearpage
\section{Experiment}
\subsection{Experimental Setting}
\label{exp:setting}
We observe that the center scaling and cropping 
operation in CLIP may crop out target objects. To address this issue 
and to align the similarity map with SAM's input, we adopt SAM's 
transformation strategy, scaling the longer side of the image to 
336 and padding the image for the CLIP input  336$\times$336. 
For results on ReasonSeg, UGround-7B-LLaVA1.5 is trained on 1 NVIDIA 
A100 GPU 80GB for about 2 days, UGround-13B-LLaVA1.5 for about 13 hours. 
The training relies solely on ReasonSeg and a referring segmentation dataset ($\sim$10k images). 
We use LoRA~\citep{iclr2022lora} for efficient fine-tuning, 
using \( lora\_r = 8 \) for 7B and 64 for 13B. 
We use the AdamW~\citep{loshchilov2017decoupled} optimizer with an initial 
learning rate of 3e-4, scheduled by WarmupDecayLR with 100 warmup steps.
Loss weights are set as $\lambda_{mask}\!=\!\lambda_{txt}\!=\!1.0$, $\lambda_{policy}\!=\!0.1$, $\lambda_{bce}\!=\!2.0$, and $\lambda_{dice}\!=\!4.0$.
\subsection{Comparison with the State-of-the-Art}
\textbf{Results on ReasonSeg Dataset.} 
In Table~\ref{table:reason_seg}, UGround achieves competitive performance compared with the recent state-of-the-art RSVP-GPT. On the ReasonSeg validation set, UGround-7B improves over RSVP-GPT by +1.4\% gIoU and +9.0\% cIoU. On the test set, UGround surpasses RSVP-GPT by +3.3\% gIoU and +5.4\% cIoU. 
At the larger scale, UGround-13B further pushes the state of the art, 
reaching 67.9\% gIoU / 74.9\% cIoU on val and 65.0\% gIoU / 65.5\% cIoU on test, outperforms READ-13B by up to +2.8\% gIoU and +2.7\% cIoU. These results validate UGround’s effectiveness in complex reasoning-oriented segmentation.
\textbf{Results on RefCOCO(+/g) Dataset.}  In Table~\ref{table:refer_seg}, UGround-7B achieves stronger overall performance than the recent state-of-the-art GLaMM-7B. 
On RefCOCO, UGround surpasses GLaMM-7B by +1.1\% on val and +0.8\% on testB. On RefCOCO+, it exceeds GLaMM-7B by +1.0\% on testB and remains competitive on val and testA. On RefCOCOg, UGround reaches 76.1\% on test(U), improving +1.2\% over GLaMM-7B, highlighting its generalization across classical referring expression segmentation benchmarks.
\textbf{Results on gRefCOCO Dataset.} 
In Table ~\ref{tab:gres}, UGround-7B achieves substantially stronger performance compared with the recent state-of-the-art GSVA-7B (ft) on the gRefCOCO dataset. On the validation set, UGround improves over GSVA-7B (ft) by +5.99\% gIoU, +2.27\% cIoU, and +12.10\% N-acc, respectively. On Test A, UGround  surpasses GSVA-7B (ft) by +3.21\% gIoU, +0.94\% cIoU, and +8.62\% N-acc. On Test B, UGround also outperforms the previous SOTA across all metrics.
These results demonstrate that UGround not only reflects its strength in multi-object segmentation but also excels in correctly identifying empty targets (N-acc).
\subsection{Ablation Study}
\label{exp:ablation}
In this section, we conduct an ablation study to analyze the contribution of each component. We report the gIoU and cIoU performance on the \textit{val} set of ReasonSeg dataset.

\textbf{Effect of the components of PPM.} 
In Table~\ref{table:ppm}, comparing Exp. 3, Exp. 6, we observe that dynamic layer selection yields a clear improvement (+2.07\% gIoU and +5.02\% cIoU
), highlighting the benefit of leveraging information across multiple layers. Moreover, the decomposition indicates that using the similarity map as a prompt ($\mathcal{M}$\textsubscript{prompt}) contributes the most (\eg, Exp. 1 vs. Exp. 2, and Exp. 4 vs. Exp. 5) within this paradigm.
\begin{table}[htbp]
    \centering
    \caption{Ablation study on PPM.}
    \label{table:ppm}
    \vspace{-6pt}
    \resizebox{1\linewidth}{!}
    {
    \tabcolsep=0.3cm
        \begin{tabular}{c|ccc|cc}
        \toprule
        Exp. ID &  SSC &
        \texttt{<SEG>}\textsubscript{prompt}
        &$\mathcal{M}$\textsubscript{prompt} & gIoU & cIoU \\
        \midrule
        1 &  & \Checkmark & & 19.13 & 15.62 
        \\
        2 &  & &\Checkmark& 60.30 & 53.86 
        \\
        3 &  &\Checkmark &\Checkmark& 64.06 & 67.05 
        \\
        \midrule
        4 & \Checkmark & \Checkmark & & 31.76 & 34.08  \\
        5 & \Checkmark & &\Checkmark & 52.14 & 52.42  \\
        6 & \Checkmark & \Checkmark & \Checkmark &\textbf{66.13} & \textbf{72.07} \\
        \bottomrule
    \end{tabular}
    }
    \vspace{-0.4cm}
\end{table}%

\textbf{Effect of explicit constraints for similarity map $\mathcal{M}$.}
In Table~\ref{table:explicit_constraints}, Exp. 1, 2, and 5 optimize the similarity map against hard labels (binary masks), while Exp. 3, 4, and 6 optimize it against soft labels (continuous masks). The results demonstrate that converting the ground-truth mask into a Gaussian heatmap smooths the object boundaries, 
leading to performance gains of +2.21\% gIoU and +3.52\% cIoU when comparing Exp. 5 and Exp. 6.
\begin{table}[htbp]
    \centering
    \caption{Ablation study on loss constraints  for similarity map $\mathcal{M}$.}
    \label{table:explicit_constraints}
    \vspace{-7pt}
    \resizebox{1\linewidth}{!}
    {
     \tabcolsep=0.1cm
        \begin{tabular}{c|cccc|cc}
        \toprule
        Exp. ID &  BCE & Dice & Gaussian BCE & Gaussian Dice & gIoU & cIoU \\
        \midrule
        1 & \Checkmark & & & & 64.56 & 71.44 \\
        2 & & \Checkmark & & & 64.57 & 68.63  \\
        3 & & & \Checkmark & & 63.16 & 65.74 \\
        4 & & & & \Checkmark & 65.80 & \textbf{73.49} \\
        5 & \Checkmark & \Checkmark & & & 63.92 & 68.55 \\
        6 & & & \Checkmark & \Checkmark & \textbf{66.13} & 72.07 \\
        \bottomrule
        \end{tabular}
    }
    \vspace{-0.4cm}
\end{table}%

\textbf{Effect of policy reward components.}
We use the negative loss of the similarity map relative to the ground-truth mask as the reward. During training, it is treated as a scalar and detached from the computation graph. In Table~\ref{table:policy_reward}, comparing Exp. 1 and Exp. 2, we observe that using $r_{bce}$ as the reward signal outperforms using $r_{dice}$, yielding a gain of +2.03\% gIoU and +1.20\% cIoU.
\begin{table}[htbp]
    \centering
    \caption{Ablation study on reward.}
    \label{table:policy_reward}
    \vspace{-7pt}
    \resizebox{1\linewidth}{!}
    {
    \tabcolsep=0.6cm
    \begin{tabular}{c|cc|cc}
        \toprule
        Exp. ID & $r_{bce}$ & $r_{dice}$ & gIoU & cIoU \\
        \midrule
        1 & \Checkmark & & 65.37 & 69.53 \\
        2 &  & \Checkmark& 63.34 & 68.33 \\
        3 & \Checkmark & \Checkmark &\textbf{66.13} & \textbf{72.07} \\
        \bottomrule
    \end{tabular}
    }
\end{table}%

\begin{figure*}[htbp]
    \begin{center}
        \includegraphics[width=0.96\linewidth]{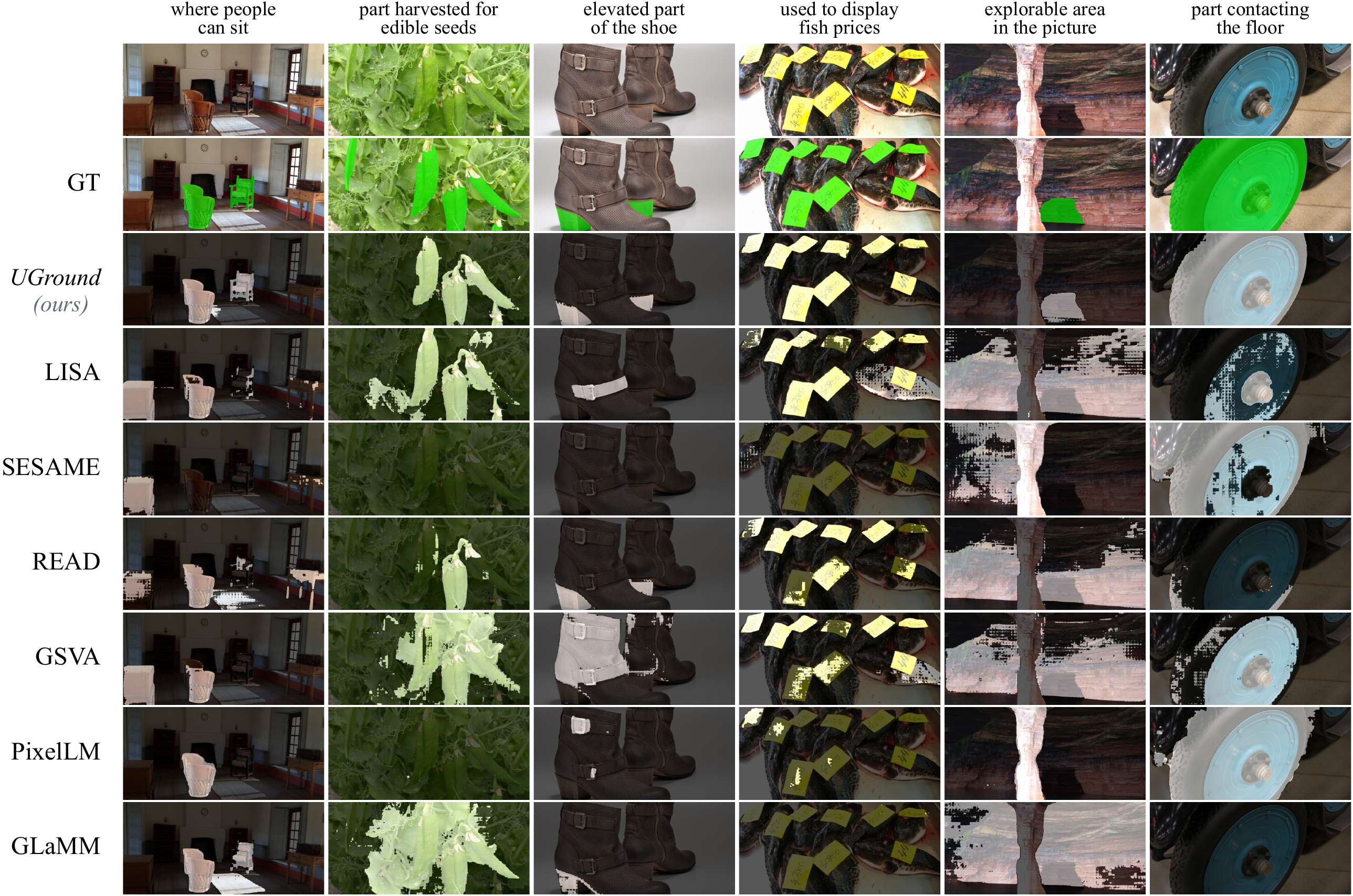}
    \end{center}
    \vspace{-0.3cm}
    \caption{ Visual comparison of UGround (Ours) with prior works on the ReasonSeg
    \textit{test} set. }
\label{fig:showcase_comparison_uground}
    \vspace{-0.42cm}
\end{figure*}

\subsection{Qualitative Results}
\label{sec:vis}
In Figure~\ref{fig:showcase_comparison_uground}, UGround generates more accurate and semantically consistent segmentation masks than existing methods, including LISA~\cite{lai2024lisa}, SESAME~\cite{wu2024see}, READ~\cite{qian2024reasoning}, GSVA~\cite{xia2024gsva}, PixelLM~\cite{ren2024pixellm}, and GLaMM~\cite{rasheed2024glamm}. 
In challenging cases that require functional or commonsense grounding, such as ``where people can sit'' and ``part harvested for edible seeds,'' UGround accurately localizes the target regions while avoiding irrelevant background regions. 
Moreover, UGround exhibits stronger fine-grained segmentation capability for small or part-level targets, such as ``the elevated part of the shoe'' and ``the region used to display fish prices.'' 
Compared with competing methods that often produce incomplete, noisy, or over-expanded masks, UGround yields cleaner boundaries and better alignment with the ground-truth masks.

\section{Conclusion}
In this work, we first visualize all unrolled Transformer layers of LMMs, represented by LLaVA, and analyze the influence of intermediate layers on downstream task decoding. Building on this analysis, we present UGround, 
which features a dynamic layer selection and prompting mechanism—Policy-Prompted Masking (PPM). PPM enables intermediate layers to connect to the SAM decoder 
in a skip-connection-like manner and adjust its connectivity in a dropout-like fashion under the ``mask as prompt'' paradigm, together contributing to unified visual grounding. Our future work aims to further investigate how 
the \texttt{<SEG>} token interacts with the integrated SAM module for alignment.
\nocite{langley00}
\begingroup
\small
\paragraph{Acknowledgments.} The computations in this research were
performed using the CFFF platform of Fudan University.

\section*{Impact Statement} This paper presents UGround to advance a unified paradigm for visual grounding. By dynamically selecting intermediate layers in unrolled transformers and using ``mask as prompt'' to provide explicit spatial cues to a segmentation model (\eg, SAM), UGround supports a broad range of settings within one system, spanning traditional referring expression segmentation to reasoning segmentation, single-target to multi-target queries, and rejection under false-premise (empty-target). In terms of societal impact, visual grounding can benefit human–AI interaction for image understanding, data annotation, and content editing. However, they may also be misused for privacy-invasive or surveillance applications (\eg, automatically localizing and extracting sensitive regions in personal imagery), or to facilitate the creation and dissemination of misleading content (\eg, precise cutouts and compositing of specific subjects). To mitigate these risks, we adhere to established community standards for responsible deployment. In particular, we encourage auditability in real-world applications and transparency in model design to foster trust.

\par
\endgroup
\bibliography{icml2026}
\bibliographystyle{icml2026}

\newpage
\appendix
\onecolumn
\section{Notations}
\label{notation}
The notations used throughout this paper are summarized in Table~\ref{notation} for clarity.
\vspace{-0.2cm}
\begin{table}[h]
    \caption{Key notations used in this paper.}
    \vspace{-0.3cm}
    \label{notation}
    \begin{center}
\resizebox{0.7\linewidth}{!}
    {
    \tabcolsep=0.3cm
        \begin{tabular}{lp{9cm}}
        \toprule
        Notation & Description \\ \midrule
        $\mathcal{G}_{\theta}$ & LMMs-based segmentation model, including a multi-modal LLM $\mathcal{G} _{\mathcal{T}}$ and a visual backbone model $\mathcal{G} _{\mathcal{V}}$,  $\mathcal{G} _{\theta}=\mathcal{G} _{\mathcal{V}} \circ \mathcal{G} _{\mathcal{T}}$ \\
        $\mathcal{H}^{(\ell)}$    & the hidden states at the $\ell$-th transformer layer   \\  $\boldsymbol{h}_{t^*}^{(\ell )}$ & the hidden state of \texttt{<SEG>} token at layer $\ell$-th (position $t^*)$ \\
        $\hat{\mathbf{M}}$ & the predicted binary segmentation mask\\
        $M$ & the ground-truth mask\\
        $M_\sigma$ & the softened label obtained by applying a Gaussian kernel to $M$ \\
        $\varphi(\cdot)$ &  a multilayer perceptron (MLP) projection layer \\
        $\mathcal{M}$\textsubscript{prompt} & mask as a prompt for SAM decoder $\mathcal{G} _{\mathcal{V}}^{dec}$ \\
        $\texttt{<SEG>}\textsubscript{prompt}$ & \texttt{<SEG>} token as prompt for SAM decoder $\mathcal{G} _{\mathcal{V}}^{dec}$  \\
        $\mathcal{M}$\textsubscript{IoU} & the overlap between the similarity map and the ground-truth mask \\
        
        $\mathcal{S}^{(\ell)}$ & the similarity scores at the layer \(\ell\)-th, \\
        $\mathcal{M}$ & the soft logit mask or similarity map $\mathcal{M} \in [0,1]^{H \times W}$ \\
        $\pi_\theta(\cdot \mid \mathcal{H}_{t^*})$ & the policy distribution conditioned on $\mathcal{H}_{t^*}$
\\
        $G_\sigma(x, y)$ & a 2D Gaussian kernel with $\sigma$, $G_\sigma(x, y) \propto \exp(- \frac{x^2 + y^2}{2\sigma^2})$\\
        
        \bottomrule
        \end{tabular}
        }
    \end{center}
\end{table}
\vspace{-0.8cm}

\section{Analysis}

\textbf{Complexity Analysis.} To justify the complexity of RL-based layer selection (RL-gating), we compare it against a simpler non-RL alternative—soft-gating, \ie, a learned soft-attention mechanism that produces a weighted average over layer outputs. We
report results on ReasonSeg val set. In Table~\ref{table:gating}, RL-gating consistently outperforms soft-gating, with gains of +1.47\% in gIoU and +1.34\% in cIoU.
\begin{table}[htbp]
    \centering
    \caption{RL-gating vs. soft-gating.}
    \label{table:gating}
    \resizebox{0.4\linewidth}{!}
    {
    \tabcolsep=0.3cm
    \begin{tabular}{c|l|cc}
        \toprule
        Exp. ID & Method & gIoU & cIoU \\
        \midrule
        1 & soft-gating & 64.66 & 70.73 \\
        2 & RL-gating   & 66.13 & 72.07 \\
        \bottomrule
        \end{tabular}
    }
    \vspace{-0.4cm}
\end{table}%

\textbf{Runtime Analysis.} To assess the computational costs of UGround, we report training and inference latency on a single NVIDIA A100-SXM4-40GB GPU in Table~\ref{table:training_cost} and Table~\ref{table:runtime_speed}. We train on the ReasonSeg train split over 400 samples for 1 epoch (200 steps), with a batch size of 2. Training Latency (s) refers to the average time per iteration during the 200-step training process (including forward, backward, and optimizer step), and Memory Usage refers to the MaxMemAllocated observed within the 200 steps. We also report the model’s Trainable Percentage (Trainable Parameters / Total Parameters). For inference, we evaluate on the ReasonSeg test split over 799 samples, with a batch size of 1. Speed (FPS) refers to the average frames per second computed over the 799 samples. In Table ~\ref{table:training_cost}, UGround exhibits a higher training latency (2.97s) than LISA~\citep{lai2024lisa}(1.26s) due to the explicit optimization of the similarity map during the backward pass. In Table ~\ref{table:runtime_speed}, UGround achieves a comparable inference runtime to LISA~\citep{lai2024lisa} and SESAME~\citep{wu2024see}.
\begin{table}[htbp]
\centering
\caption{Comparing the training cost of our UGround to state-of-the-art methods.}
\label{table:training_cost}
\resizebox{1\linewidth}{!}
    {
    \tabcolsep=0.3cm
\begin{tabular}{lccccc}
\toprule
Model & Training Latency (s) & Memory Usage (GB) & Trainable (\%) & Trainable Params & Total Params \\
\midrule
SESAME~\citep{wu2024see} & 1.11 & 23.15 & 3.73\% & 288.25M & 7.73B \\
PixelLM~~\citep{ren2024pixellm}    & 1.22 & 23.02 & 5.25\% & 375.72M & 7.16B \\
LISA~\citep{lai2024lisa}   & 1.26 & 23.68 & 3.74\% & 288.25M & 7.71B \\
GSVA~\citep{xia2024gsva}       & 2.13 & 25.73 & 3.73\% & 288.26M & 7.73B \\
UGround    & 2.97 & 28.16 & 3.72\% & 288.40M & 7.75B \\
\bottomrule
\end{tabular}
}
\vspace{-0.2cm}
\end{table}
\begin{table}[htbp]
\centering
\caption{Comparing the runtime speed of our UGround to state-of-the-art methods.}
\vspace{-0.2cm}
\label{table:runtime_speed}
\resizebox{1\linewidth}{!}
    {
    \tabcolsep=0.25cm
\begin{tabular}{lccccc}
\toprule
Model & GSVA~\citep{xia2024gsva} & SESAME~\citep{wu2024see} & LISA~\citep{lai2024lisa} & PixelLM~\citep{ren2024pixellm} & UGround \\
\midrule
Speed (FPS) & 3.98 & 4.64 & 4.68 & 9.24 & 4.12 \\
\bottomrule
\end{tabular}
}
\vspace{-0.5cm}
\end{table}

\textbf{Generalization Analysis.} To demonstrate broader generalization of UGround, we report results on MUSE benchmark. MUSE aims for multi-target reasoning segmentation~\citep{ren2024pixellm}. In Table~\ref{table:multireason_seg}, UGround-7B consistently outperforms PixelLM-7B across all splits of the MUSE benchmark. It improves the Val overall cIoU by +5.7\% and boosts the challenging “few-targets” cIoU by +10.5\%. On the Test overall set, UGround also achieves a +4.1\% cIoU gain, yielding a clear performance margin over PixelLM-7B. Also, we report results on augmented FP-Refcoco(+/g) under false premises. In Table~\ref{table:refer_seg_see_segment}, UGround outperforms READ across all FP-Refcoco(+/g) splits, with gains of +1.59\% to +2.99\% in ``see'' accuracy and +1.30\% to +2.43\% in segment scores (cIoU).
\begin{table}[htbp]
    \centering
    \caption{Comparison on MUSE benchmark. Numbers are cited from PixelLM~\citep{ren2024pixellm}.}
    \vspace{-0.2cm}
    \label{table:multireason_seg} 
    \resizebox{1\linewidth}{!}
    {
    \tabcolsep=0.53cm
        \begin{tabular}{l|c|cc|cc|cc|cc }
            \toprule
            
            \multirow{3}*{Method} &\multirow{3}*{\makecell{w/o\\SAM}}  & \multicolumn{2}{c|}{Val} & \multicolumn{6}{c}{Test}   \\ 
            
            \specialrule{0em}{0pt}{1pt}
            \cline{3-10}
            \specialrule{0em}{0pt}{1pt}
              & & \multicolumn{2}{c|}{overall}  
              & \multicolumn{2}{c|}{few targets} 
              & \multicolumn{2}{c|}{many targets} 
              & \multicolumn{2}{c}{overall}  \\ 

            \specialrule{0em}{0pt}{1pt}
            \cline{3-10}
            \specialrule{0em}{0pt}{1pt}
            
              & & gIoU & cIoU & gIoU & cIoU & gIoU & cIoU & gIoU & cIoU    \\ 
            
            \specialrule{0em}{0pt}{1pt}
            \hline
            \specialrule{0em}{0pt}{1pt}

            {\color{lightgray}SEEM}~\citep{zou2023seem}& {\color{lightgray}\Checkmark}   
            &{\color{lightgray}13.6}  &{\color{lightgray}16.2} 
            &{\color{lightgray}23.6} &{\color{lightgray}24.9} 
            &{\color{lightgray}8.5} &{\color{lightgray}13.2}  
            &{\color{lightgray}11.7} &{\color{lightgray}15.7} \\
            
            \specialrule{0em}{0pt}{1pt}
            \hline
            \specialrule{0em}{0pt}{1pt}
            
            LISA-7B~\citep{lai2024lisa} &  \XSolidBrush 
            & 18.8 &29.0  
            &24.7 &36.5 
            &9.6 &24.5  
            &12.8  &27.1   \\

            LISA-7B$_{\text{rec}}$~\citep{lai2024lisa} &  \XSolidBrush 
            &24.5  &31.1 
            &30.0 &30.9 
            &12.4  &23.2  
            &16.2 &24.8  \\
            
            LISA-7B$_{\text{aug}}$~\citep{lai2024lisa} &  \XSolidBrush 
            &42.0 &46.1  
            &43.5  &52.0 
            &37.7 &42.3  
            &38.9 &44.4 \\

            PixelLM-7B~\citep{ren2024pixellm} &  \Checkmark 
            &42.6  &50.7  
            &44.6 &59.2  
            &37.7 &42.8  
            &39.2 &46.3\\

            UGround-7B &  \Checkmark 
            &\textbf{52.4}  &\textbf{56.4}  
            &\textbf{46.2} &\textbf{69.7}  
            &\textbf{38.8} &\textbf{46.9}  
            &\textbf{40.1} &\textbf{50.4}\\

            \bottomrule            
        \end{tabular}
    }
\vspace{-0.3cm}
\end{table}
\begin{table}[htbp]
    \centering
    \vspace{-0.1cm}
    \caption{
    Comparisons of the state-of-the-art ``see'' and ``segment'' results on augmented FP-Refcoco(+/g) \textit{val} set.
    ``FP''(False Premise) denotes a query for an object absent from the image.}
    \vspace{-0.2cm}
\resizebox{1\linewidth}{!}
    {
    \tabcolsep=0.7cm
        \begin{tabular}{ l | c c | c c | c c }
            \toprule
            
            \multirow{2}*{Method} & \multicolumn{2}{c|}{FP-RefCOCO} & \multicolumn{2}{c|}{FP-RefCOCO+}  & \multicolumn{2}{c}{FP-RefCOCOg} \\ 
            
            \specialrule{0em}{0pt}{1pt}
            \cline{2-7}
            \specialrule{0em}{0pt}{1pt}
            ~ & See & Segment & See & Segment & See & Segment \\ 
            \specialrule{0em}{0pt}{1pt}
            \hline
            \specialrule{0em}{0pt}{1pt}
            LISA-7B \citep{lai2024lisa} & 51.36 & 44.00 & 51.32 &  39.62 & 51.25 & 39.64 \\
            Cascading \citep{wu2024see} & 75.59 &  55.18 & 75.03 & 48.64 & 76.07 & 49.98 \\
            SESAME \citep{wu2024see} & {79.84} & {57.93} & {80.00} & {50.81} & {81.78} & {53.79} \\
            READ \citep{qian2024reasoning} & 
            {82.87} & {61.50} & {83.51} & {54.54} & {84.67} & {56.12} \\
            \specialrule{0em}{0pt}{1pt}
            \hline
            \specialrule{0em}{0pt}{1pt}
            \toolname & \textbf{85.86} & \textbf{62.80} & \textbf{85.10} & \textbf{56.03} & \textbf{86.86} & \textbf{58.55}\\
            \bottomrule            
        \end{tabular}
    }
    \label{table:refer_seg_see_segment}  
    \vspace{-0.5cm}
\end{table}

\section{Discussions}
\subsection{Justification for Reinforcement Learning (RL)}
Reinforcement Learning (RL) is introduced to address the hard-boundary problem in layer selection, enabling the model to explicitly choose a layer (\eg, layer 26) in a soft, differentiable manner during training. Because hard-boundary layer selection is non-differentiable, using RL is necessary.

An intuitive approach is to train a classifier for dynamic layer selection using cross-entropy loss, followed by soft-gating via a weighted average over all layer outputs. We call this simpler alternative CE + soft-gating. Here, the layer-selection label is assigned based on the overlap (IoU) between each layer’s similarity map (layers 0–31) and the ground-truth mask. The classifier is trained, and its output scores serve as soft gates to weight and fuse the transformer layers (0–31) into a unified representation, which is then fed into SAM for mask decoding. However, CE + soft-gating suffers from the following issues:

1) CE + soft-gating is numerically suboptimal. Consider a scenario after training for 1–2 epochs where the classifier predicts the following scores for layers 0–31: ($[0.01, 0.004, \ldots, 0.7, 0.1, 0.1]$).  Since 0.7 is the maximum score, the 29-th layer is identified as the best layer.  However, if we use the full score vector ($[0.01, 0.004, \ldots, 0.7, 0.1, 0.1]$) as soft-gating weights to fuse the embeddings of layers 0–31, the best embedding from layer 29 will inevitably be mixed with features from all other layers (0–28 and 30–31), making it no longer optimal. Soft-gating inherently prevents the remaining layers from receiving zero weight.  In fact, as shown in Fig.~\ref{fig:layers1-40} ( and also discussed in~\citep{zhang2024redundancy}), transformer layers are activated only at a few specific depths. Assigning non-zero weights to inactive layers dilutes the contribution of these key activated layers. Therefore, a hard boundary is required—that is, explicitly selecting a single layer (\eg, assigning 100\% weight to layer 29) rather than distributing fractional weights (\eg, 60\% on layer 29). This is inherently a discrete decision problem, for which reinforcement learning (RL) provides a principled and trainable solution.

2) The training objective of the layer-selection classifier (\ie, CE loss) does not align with the ultimate goal of optimizing the gates through backpropagation from the downstream SAM mask decoding (\ie, use soft-gating without an explicit loss on the gates).  The layer-selection classifier explicitly trains the gate to predict which layer is the most important, whereas using soft weights to fuse all layers aims to optimize the averaged representation, without explicitly enforcing the gate to select the optimal layer. Besides, soft-gating presumably  ignores inter-layer dependencies in the transformer. By weighting and summing all layers, it actually performs a soft ensemble, disrupting the original layer-by-layer structure of the transformer.

3) Computational cost: CE + soft gating vs. RL-gating.  Given $\boldsymbol{h}_{seg} \in \mathbb{R}^{4096}$ and $\boldsymbol{h}_{img} \in \mathbb{R}^{576 \times 4096}$, the similarity map is computed as:  $\boldsymbol{h}_{seg} \times \boldsymbol{h}_{img}^{T} = (1 \times 4096) \times (576 \times 4096)^{T} = 576 $, followed by reshaping into $24 \times 24$, and then interpolating to $256 \times 256$ to obtain $\mathcal{M}$.  If using the CE + soft-gating approach, one must compute the similarity map for all $32$ layers of LLaVA, resulting in $32 \times 256 \times 256$ maps, and then compute the overlap (IoU) with the ground-truth mask as the supervision label.  In contrast, with RL, we compute $\boldsymbol{h}_{seg} \times \boldsymbol{w}^{T} = (1 \times 4096) \times (32 \times 4096)^{T} = 32$, yielding a 32-dimensional similarity logits vector, followed by REINFORCE to select the layer index, \eg, ($\ell^* = 28$). $\boldsymbol{w}^T \in \mathbb{R}^{32 \times 4096}$ denotes the layer-gating weights, which are the only learnable parameters in UGround. The reward $r$ only requires computing the similarity map for a single layer.

\subsection{Costs of Mask as Prompt}

Compared to the \texttt{<SEG>} token, Mask as Prompt (MasP) is more effective, as it provides explicit spatial cues via its activation regions.
Here, we analyze the computational costs associated with Mask as Prompt. We use SAM~\citep{kirillov2023segment} for mask decoding. The original paper describes its prompt encoder as follows:

\textit{``Prompt encoder. We consider two sets of prompts: sparse (points, boxes, text) and dense (masks). We represent points and boxes by positional encodings summed with learned embeddings for each prompt type, and free-form text with an off-the-shelf text encoder from CLIP~\citep{radford2021learning}. Dense prompts (\eg, masks) are embedded using convolutions and summed element-wise with the image embedding."}
\begin{table}[htbp]
\vspace{-0.2cm}
\centering
\caption{Mask prompt dimension changes through convolutional layers in SAM.}
\vspace{-0.1cm}
\label{tab:mask_conv}
\begin{tabular}{l|l}
\toprule
Layer & Output shape \\
\midrule
Input mask: $256\times256$ & (B, 1, 256, 256) \\
Conv2d(1 $\rightarrow$ 4, stride=2) & (B, 4, 128, 128) \\
Conv2d(4 $\rightarrow$ 16, stride=2) & (B, 16, 64, 64) \\
Conv2d(16 $\rightarrow$ 256, stride=1) & (B, 256, 64, 64) \\
\bottomrule
\end{tabular}
\vspace{-0.2cm}
\end{table}

That is, SAM accepts a mask prompt of size $256\times256$, which is transformed by the convolutional layers in Table~\ref{tab:mask_conv}.
Thus, the final mask embedding has dimensions [B, 256, 64, 64]. In SAM’s implementation, even when the mask-as-prompt input is not provided, \ie, when only the text \texttt{<SEG>} prompt is used (\texttt{mask=None}), SAM still generates a set of learnable embeddings of the same shape as a placeholder, serving as the mask-embedding representation. Therefore, using Mask as Prompt does not introduce any additional parameter overhead.

\subsection{Analogous to Monte Carlo Dropout}

Analogous to Monte Carlo Dropout. Consider the query: ``Can you segment the apple, orange, and banana in the picture ?'' UGround responds: ``Yes, apple, orange, and banana are \texttt{<SEG>}, \texttt{<SEG>}, \texttt{<SEG>}.'' That is, there are three objects to segment in a single image. During one forward pass, our SSC module (RL) selects a layer from 0–31 for each of $\texttt{<SEG>}_{apple}$, $\texttt{<SEG>}_{orange}$, and $\texttt{<SEG>}_{banana}$ as input to SAM. Suppose: in the ($1$)-th forward pass, layers (15, 28, 19) are selected, that is, $\texttt{<SEG>}_{apple}^{(15)} \rightarrow$ SAM, $\texttt{<SEG>}_{orange}^{(28)} \rightarrow$ SAM, $\texttt{<SEG>}_{banana}^{(19)} \rightarrow$ SAM; in the ($\mathcal{T}$-1)-th forward pass, the selections are $\texttt{<SEG>}_{apple}^{(16)} \rightarrow$ SAM, $\texttt{<SEG>}_{orange}^{(20)} \rightarrow$ SAM, $\texttt{<SEG>}_{banana}^{(14)} \rightarrow$ SAM; in the $\mathcal{T}$-th forward pass, the selections are $\texttt{<SEG>}_{apple}^{(13)} \rightarrow$ SAM, $\texttt{<SEG>}_{orange}^{(17)} \rightarrow$ SAM, $\texttt{<SEG>}_{banana}^{(26)} \rightarrow$ SAM, then during the $\mathcal{T}$ forward passes, $\texttt{<SEG>}_{apple}$ from layers 15, 16, and 13 are independently connected to SAM, and the SSC module dynamically alters the model’s internal connectivity. This can be analogized to the dropout mechanism in neural networks: All layers are virtually connected to SAM, but only one path is activated per forward pass. Since the SSC module performs stochastic sampling, this can be interpreted as a Monte Carlo uncertainty estimation~\citep{gal2016dropout}. Unlike soft-gating, the SSC module enables each layer to connect independently to SAM, resulting in 32 independent subnetworks in an ensemble-style design. As only one layer is activated at a time, the computational overhead is comparable to selecting a fixed last layer, thus avoiding expensive computational costs.

\section{Qualitative Results}

\begin{figure}[htbp]
    \centering
    \scalebox{1}{
    \begin{subfigure}[b]{0.43\textwidth}
        \centering        \includegraphics[width=\textwidth]{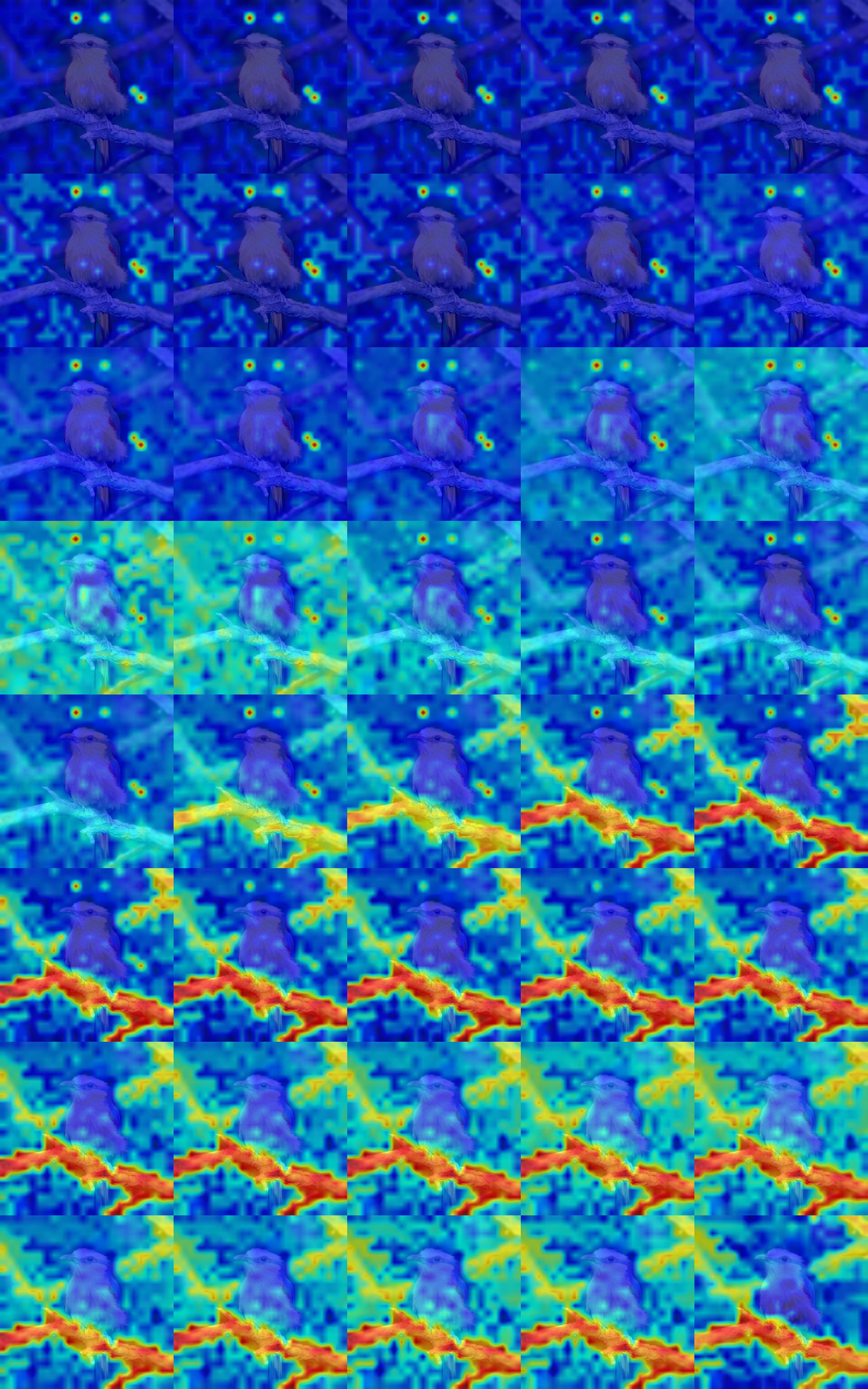}
        \caption{Fixed last hidden layer (Prior Works).}
        \label{fig:fixed_bird}
    \end{subfigure}
    \hfill 
    \begin{subfigure}[b]{0.43\textwidth}
        \centering
        \includegraphics[width=\textwidth]{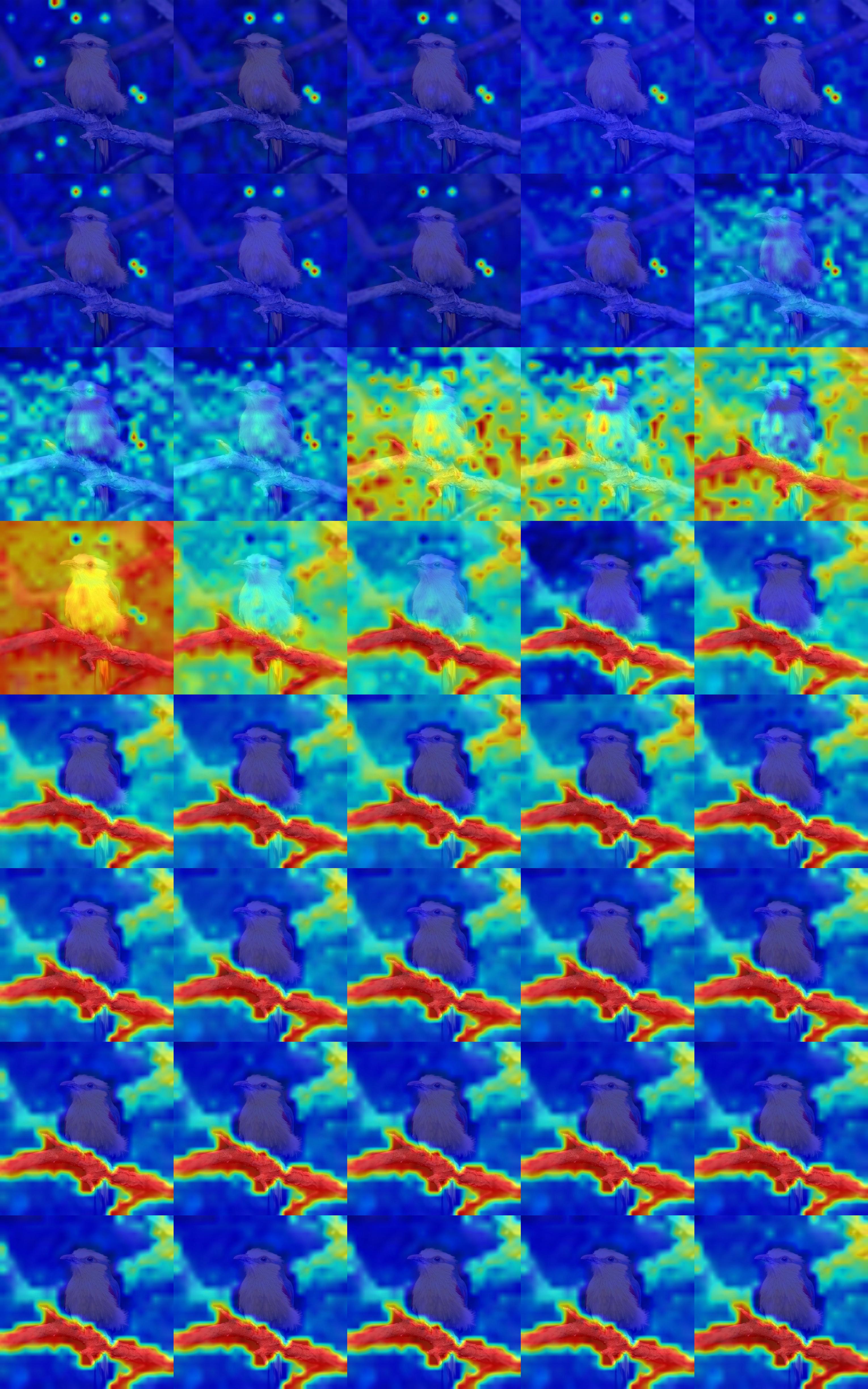}
        \caption{Policy-Prompted Masking (Ours).}
        \label{fig:dynamic_bird}
    \end{subfigure}
    }
    \caption{Similarity map across layers 1-40.}
    \label{fig:layers1-40}
    \vspace{-0.5cm}
\end{figure}

\begin{figure}[htbp]
    \vspace{-0.5cm}
    \centering
    \scalebox{0.93}{
    \begin{subfigure}[b]{0.4\textwidth}
        \centering
        \includegraphics[width=\textwidth]{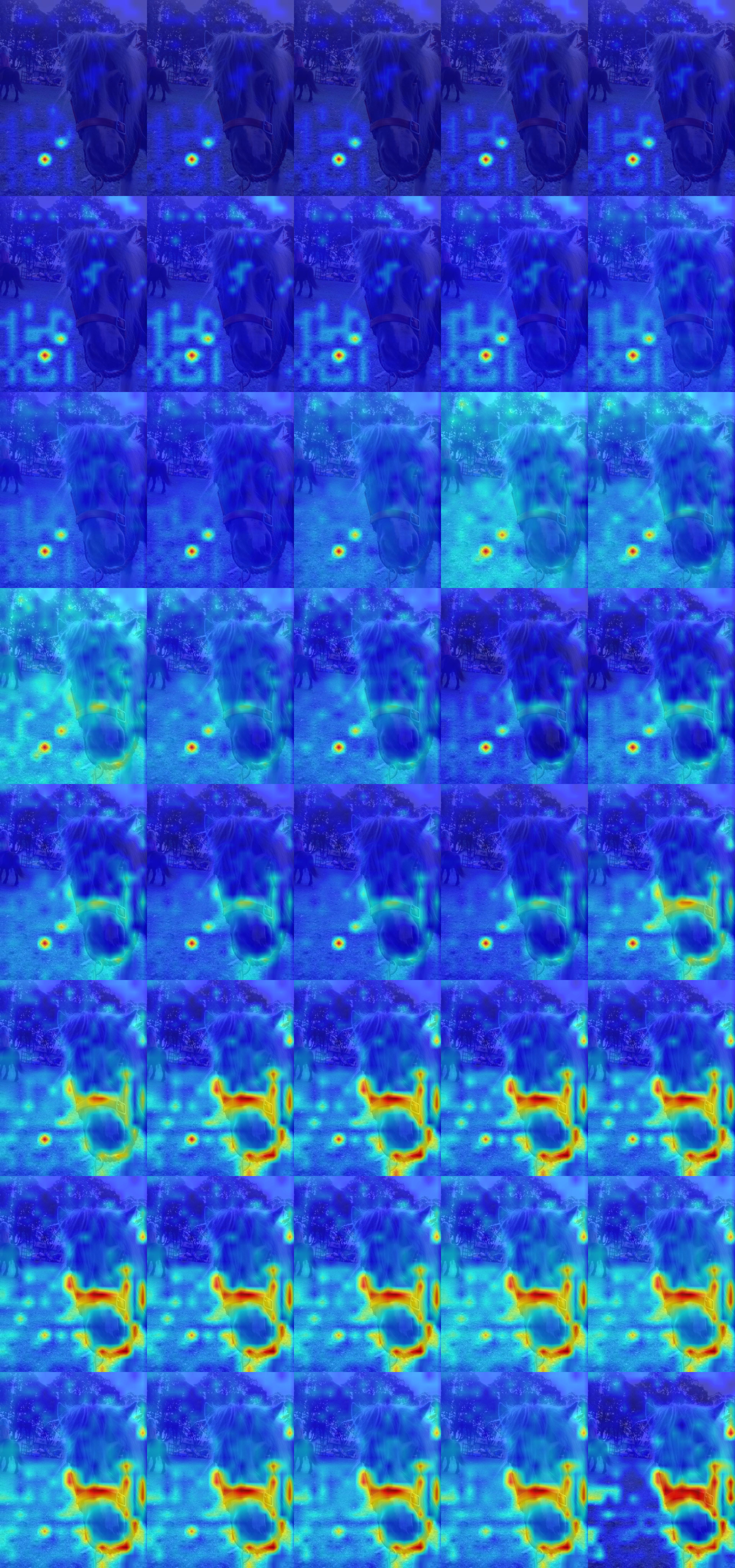}
        \caption{Fixed last layer (Prior Works).}
        \label{fig:fixed_horse}
    \end{subfigure}
    \hfill 
    \begin{subfigure}[b]{0.4\textwidth}
        \centering
        \includegraphics[width=\textwidth]{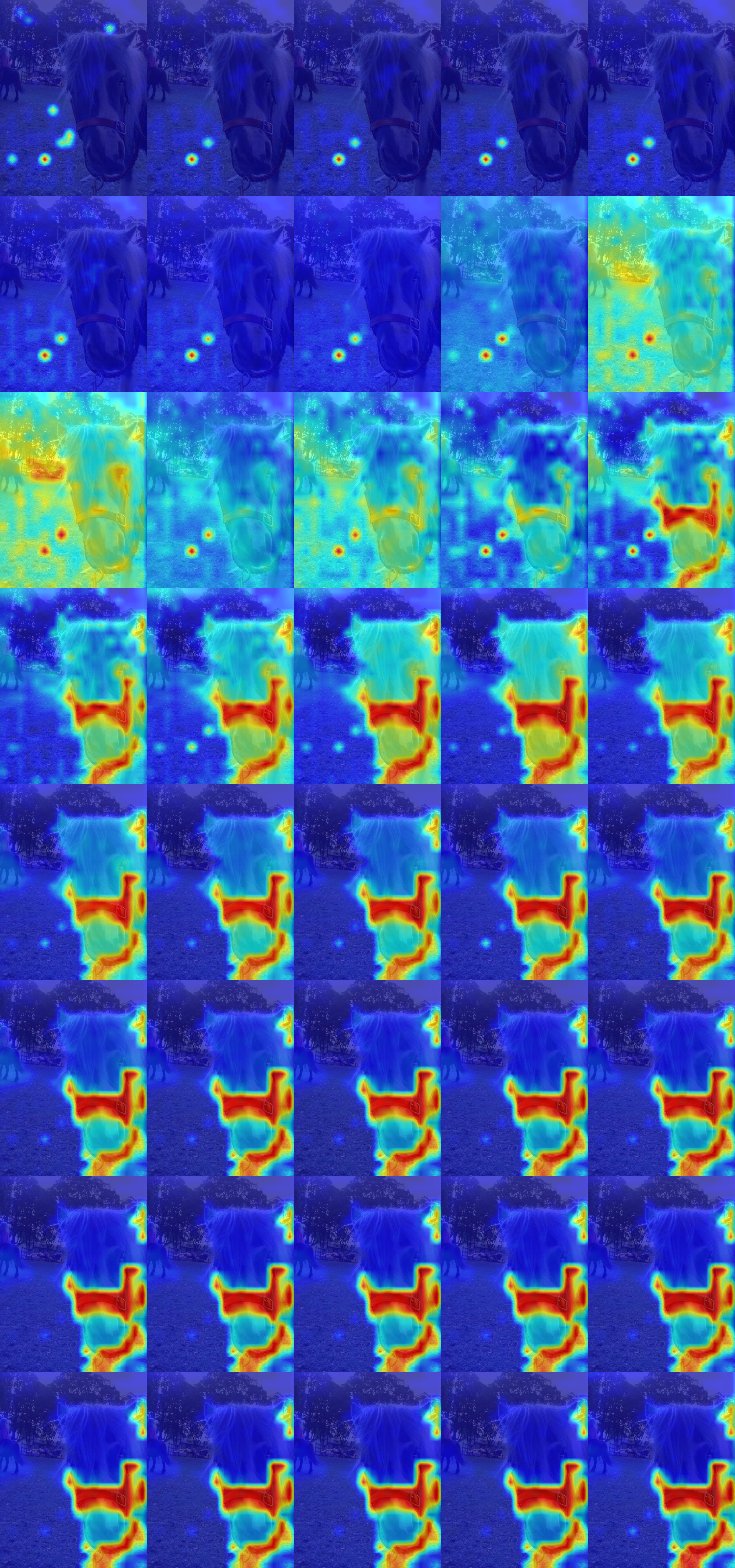}
        \caption{Policy-Prompted Masking (Ours).}
        \label{fig:dynamic_horse}
    \end{subfigure}
    }
    \caption{Similarity map across layers 1-40.}
    \label{fig:layers2}
    \vspace{-0.3cm}
\end{figure}

\begin{figure}[htbp]
\vspace{-0.3cm}
    \centering
    \scalebox{0.93}{
        \begin{subfigure}[b]{0.4\textwidth}
            \centering
            \includegraphics[width=\textwidth]{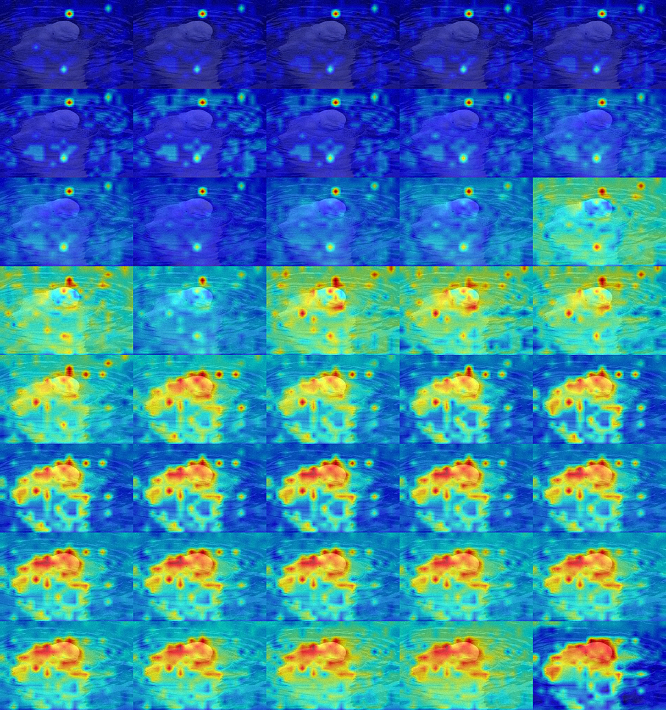}
            \caption{Fixed last hidden layer (Prior Works).}
            \label{fig:fixed_dolphin}
        \end{subfigure}
        \hfill
        \begin{subfigure}[b]{0.4\textwidth}
            \centering
            \includegraphics[width=\textwidth]{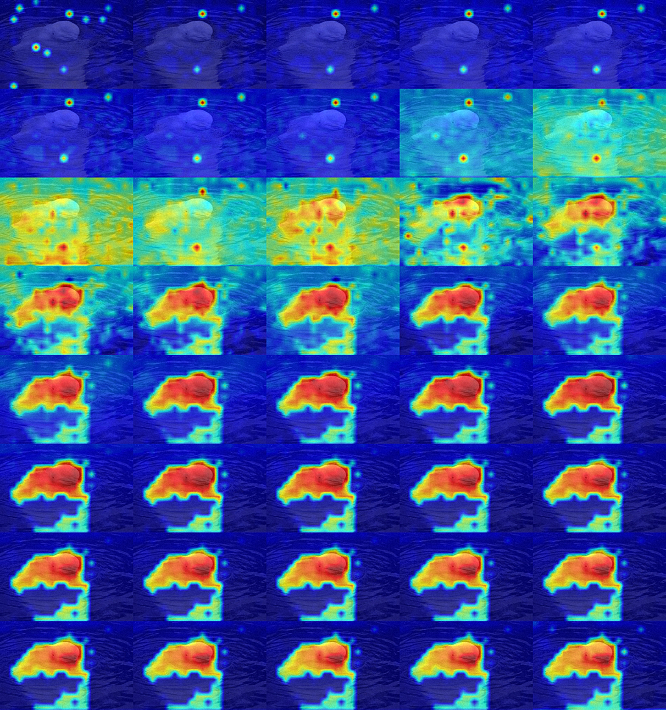}
            \caption{Policy-Prompted Masking (Ours).}
            \label{fig:dynamic_dolphin}
        \end{subfigure}
    }
    \caption{Similarity map across layers 1-40.}
    \label{fig:layers3}
\end{figure}

\textbf{Layer-wise visualization of the Transformer across layers 1-40.}
\label{app:layers1-40} Fig.~\ref{fig:layers1-40}–\ref{fig:layers3} present a qualitative analysis of similarity map activations across layers 1–40, illustrating their alignment with the ground-truth mask. Each figure contrasts heatmaps generated by prior works using a fixed last hidden layer (a) with those from our proposed Policy-Prompted Masking method (b). Unlike the baseline approach (a), our method introduces stochastic layer selection: in each forward pass, a layer $\ell^*$ is randomly chosen, and multiple $\mathcal{T}$ stochastic samples result in an approximate Bayesian uncertainty estimation, akin to Monte Carlo Dropout. This sampling mechanism explores diverse pathways, alleviates over-reliance on any single trajectory, and virtually forms an ensemble of subnetworks, thereby enhancing robustness. As a result, the similarity maps in (b) show activations that are sharply concentrated on the target object in intermediate and later layers, yielding more accurate alignment with the ground-truth mask compared to the diffuse and noise-prone activations observed in (a).

\begin{figure*}[htbp]
\begin{center}
\vspace{-0.2cm}
\includegraphics[width=0.85\linewidth]{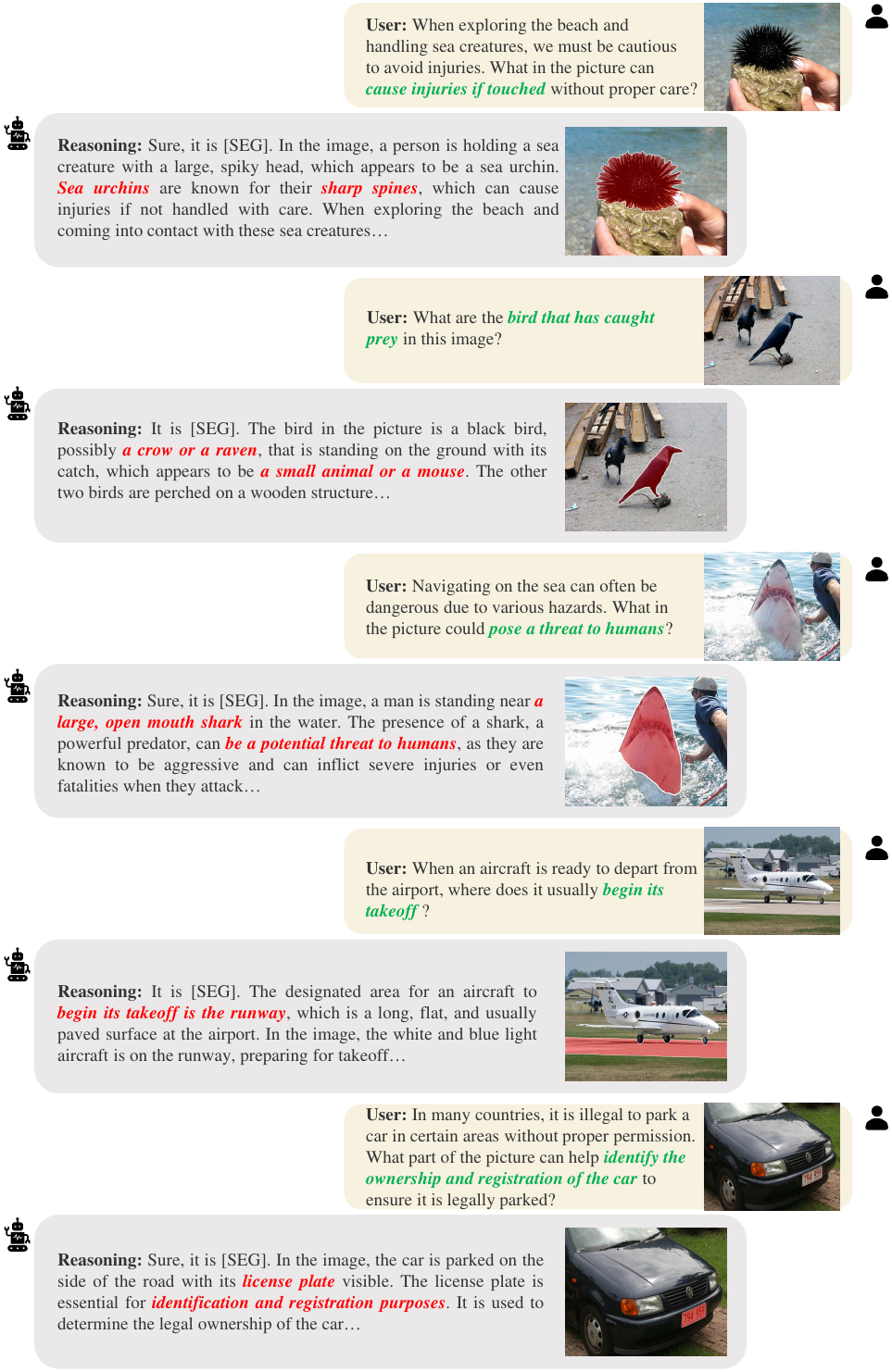}
\end{center}
\vspace{-0.35cm}
\caption{Visualization of UGround-7B on the ReasonSeg test set, demonstrating UGround’s capability in semantic segmentation, \ie, assigning pixel-level labels for a target semantic category (with background as 0). Zoom in for best view.}
\label{fig:Appendix_reasonseg_7b}
\end{figure*}

\begin{figure*}[htbp]
\begin{center}
\vspace{-0.2cm}
\includegraphics[width=0.85\linewidth]{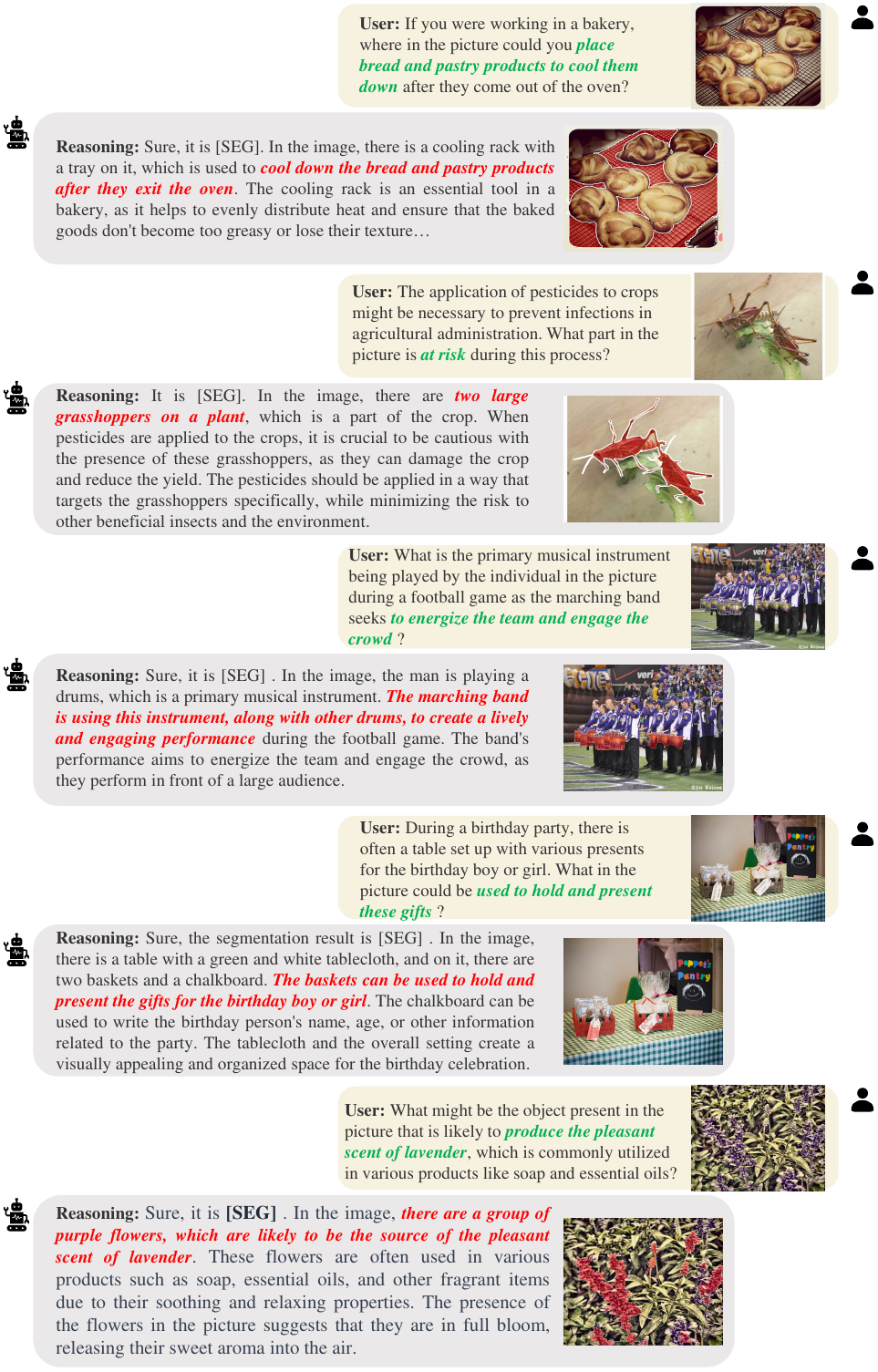}
\end{center}
\vspace{-0.35cm}
\caption{Visualization of UGround-13B on the ReasonSeg test set, demonstrating UGround’s capability in semantic segmentation, \ie, assigning pixel-level labels for a target semantic category (with background as 0). Zoom in for best view.}
\label{fig:Appendix_reasonseg_13b}
\end{figure*}

\begin{figure*}[htbp]
\begin{center}
\vspace{-0.15cm}
\includegraphics[width=0.85\linewidth]{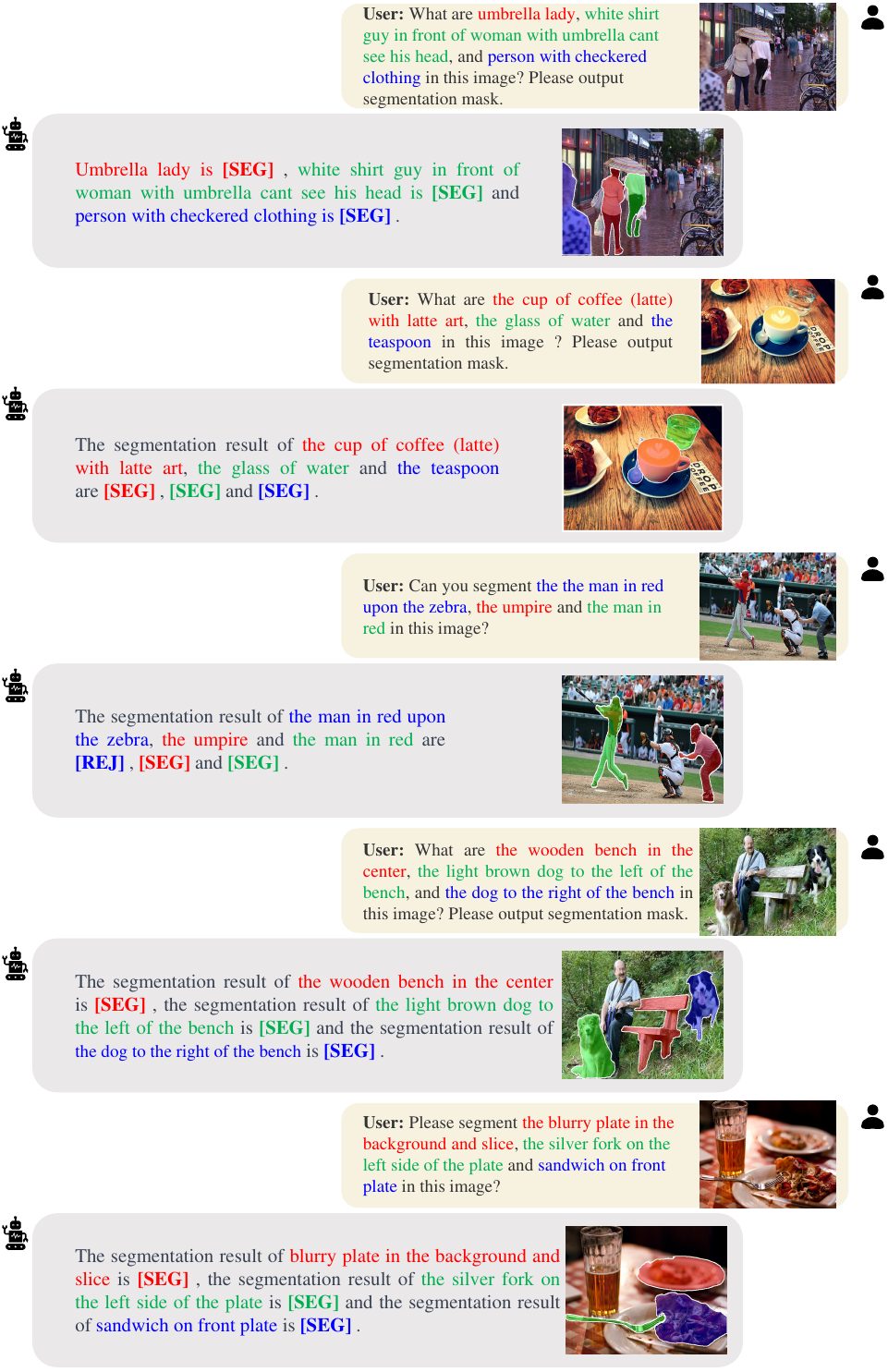}
\end{center}
\vspace{-0.35cm}
\caption{
Visualization of UGround-7B on gRefCOCO~\citep{liu2023gres} test sets A and B (I), demonstrating UGround’s capability in instance segmentation, \ie, predicting instance-level masks for individual object instances (with separate masks for objects of the same category). Zoom in for best view.}
\label{fig:Appendix_greferseg_7b_1}
\end{figure*}

\begin{figure*}[htbp]
\begin{center}
\vspace{-0.2cm}
\includegraphics[width=0.85\linewidth]{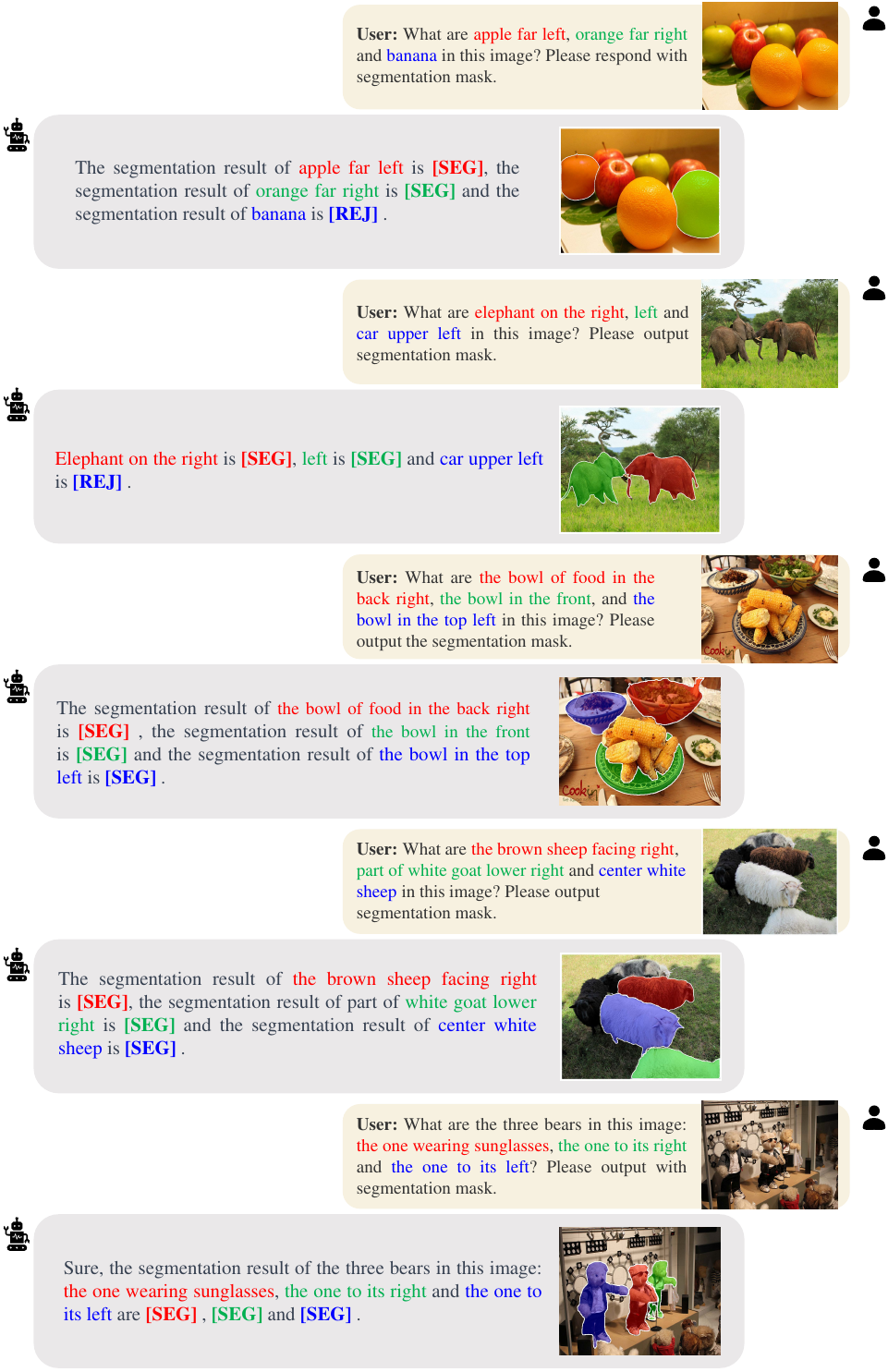}
\end{center}
\vspace{-0.35cm}
\caption{Visualization of UGround-7B on gRefCOCO~\citep{liu2023gres} test sets A and B (II), demonstrating UGround’s capability in instance segmentation, \ie, predicting instance-level masks for individual object instances (with separate masks for objects of the same category). Zoom in for best view.}
\label{fig:Appendix_greferseg_7b_2}
\end{figure*}

\begin{figure*}[h]
\begin{center}
\vspace{-0.1cm}
\includegraphics[width=0.96\linewidth]{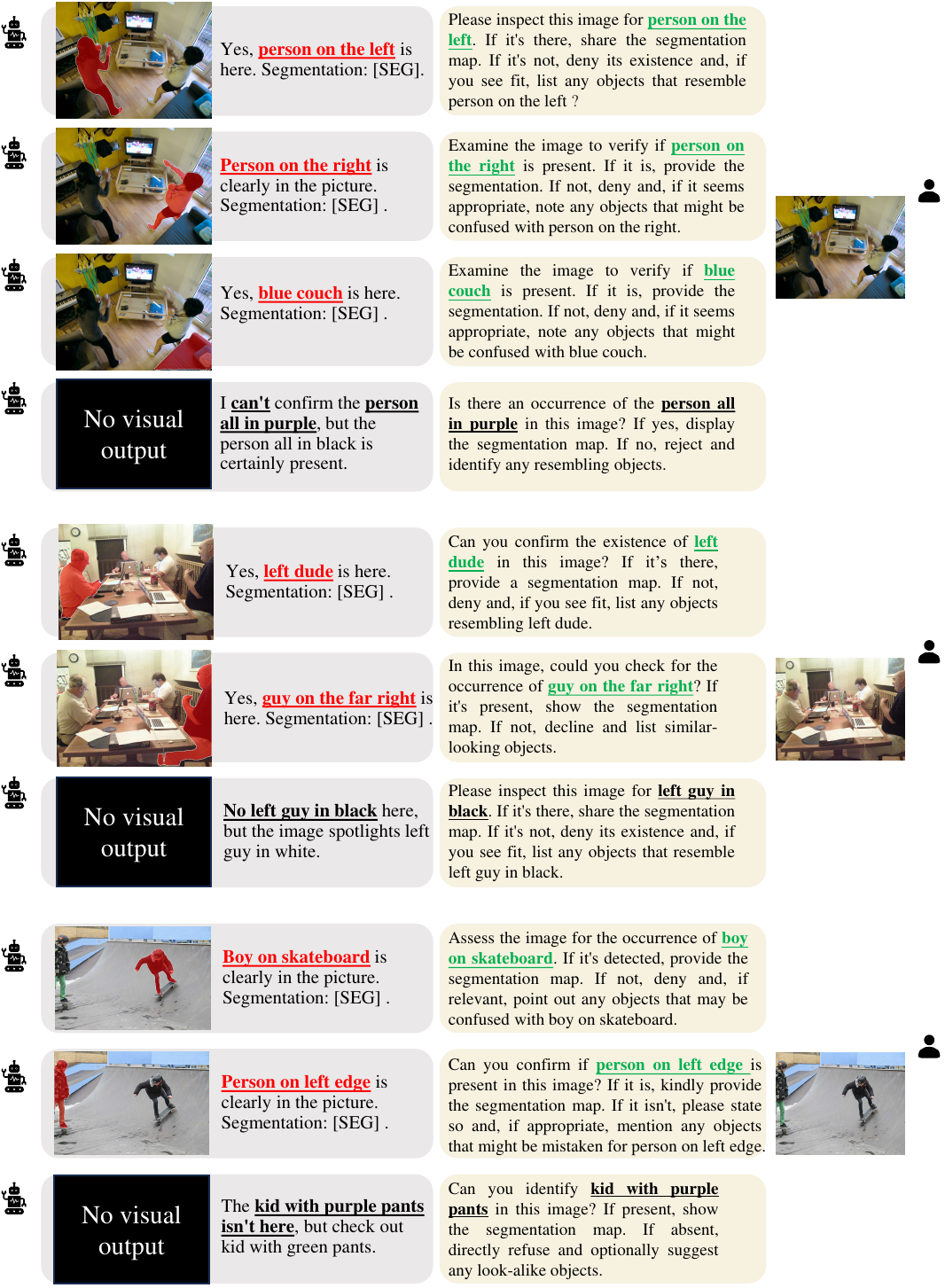}
\end{center}
\vspace{-0.2cm}
\caption{
Visualization of UGround-7B on the FP-RefCOCO(+/g) val set, demonstrating UGround’s capability in handling false-premise queries, \ie, queries that refer to objects not present in the given image (empty target). Zoom in for best view.}
\label{fig:Appendix_fpreferseg_7b}
\end{figure*}

\textbf{Visualization of UGround on benchmarks.} 
Fig.~\ref{fig:Appendix_reasonseg_7b}–\ref{fig:Appendix_reasonseg_13b} show qualitative results on ReasonSeg test set. Fig.~\ref{fig:Appendix_reasonseg_13b} demonstrates UGround’s capability in semantic segmentation, \ie, assigning pixel-level labels for a target semantic category (with background as 0). Fig.~\ref{fig:Appendix_greferseg_7b_1}–\ref{fig:Appendix_greferseg_7b_2} show qualitative results on gRefCOCO~\citep{liu2023gres} test sets A and B, which demonstrate UGround’s capability in instance segmentation, \ie, predicting instance-level masks for individual object instances (with separate masks for objects of the same category). 
Fig.~\ref{fig:Appendix_fpreferseg_7b} shows qualitative results on FP-RefCOCO(+/g) val set, which demonstrates UGround’s capability in handling false-premise queries, \ie, queries that refer to objects not present in the given image (empty target). Notably, UGround is capable of correcting false premises, \ie, directly refusing queries when the target object is absent and optionally suggesting look-alike objects~\cite{wu2024see} as shown in Fig.~\ref{fig:Appendix_fpreferseg_7b}.

\section{Implementation Details}
\textbf{The UGround code repository.} 
Currently, UGround supports 8 dataset types, namely, A: sem\_seg, B: refer\_seg, C: neg\_refer\_seg, D: correct\_refer\_seg, E: vqa, F: reason\_seg, G: reason\_seg\_plus, and H: multi\_reason\_seg.
In particular, 

{A: sem\_seg: ade20k $||$ cocostuff $||$ pascal\_part $||$ paco\_lvis $||$ mapillary}

{B: refer\_seg: refclef $||$ refcoco $||$ refcoco+ $||$ refcocog $||$ refzom $||$ grefcoco}

{C: neg\_refer\_seg: R-Refcoco $||$ R-Refcoco+ $||$ R-Refcocog}

{D: correct\_refer\_seg: fprefcoco $||$ fprefcoco+ $||$ fprefcocog}

{E: vqa: llava\_instruct\_150k}

{F: reason\_seg: ReasonSeg$|$train}

{G: reason\_seg\_plus(LISA++): instance\_seg $||$ cot $||$ conversations $||$ caption}

{H: multi\_reason\_seg(muse): MultiReasonSeg$|$train}

\begin{figure*}[h]
\begin{center}
\includegraphics[width=0.8\linewidth]{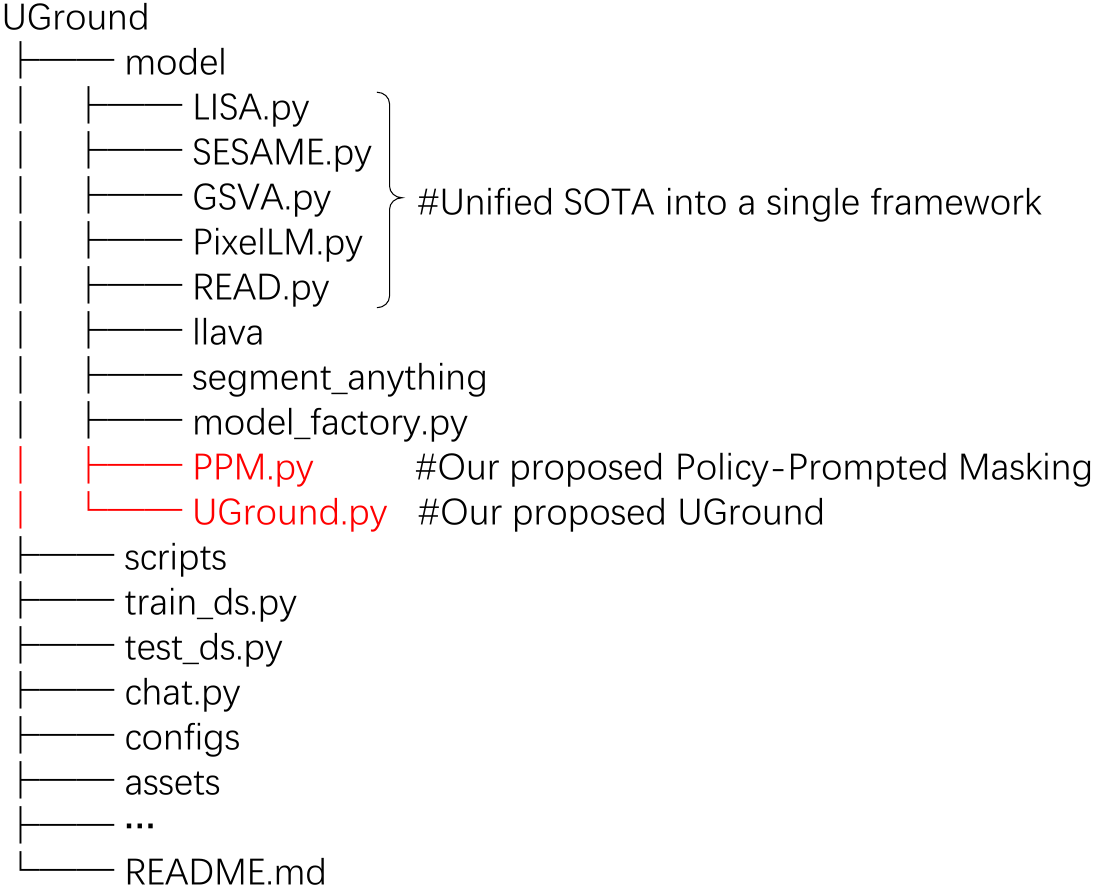}
\end{center}
\caption{The UGround code repository. All codes and models are publicly available at {\hypersetup{urlcolor=colorlink}\href{https://github.com/rui-qian/UGround}{https://github.com/rui-qian/UGround}}.}
\label{fig:dir}
\vspace{-0.2cm}
\end{figure*}

Our UGround is trained in a mixed fashion on these 8 types of ``in-the-wild" datasets as needed. As shown in Fig.~\ref{fig:dir}, the UGround project is organized with a model directory for core implementations, a scripts directory containing training and testing scripts, and a configs directory for configuration files. The main model architecture is defined in UGround.py within the UGroundForCausalLM class, which extends LlavaLlamaForCausalLM and integrates a Segment Anything Model (SAM) for mask generation. A central component of this architecture is the PolicyPromptedMasking class, detailed in PPM.py, which is instantiated and used by the UGroundForCausalLM class to dynamically select hidden state features from different network layers. The selection policy is trained using the REINFORCE algorithm with either an EMA or a critic-based baseline. 
All codes and models are publicly available at {\hypersetup{urlcolor=colorlink}\href{https://github.com/rui-qian/UGround}{https://github.com/rui-qian/UGround}}.

\begin{figure*}[h]
\begin{center}
\includegraphics[width=0.98\linewidth]{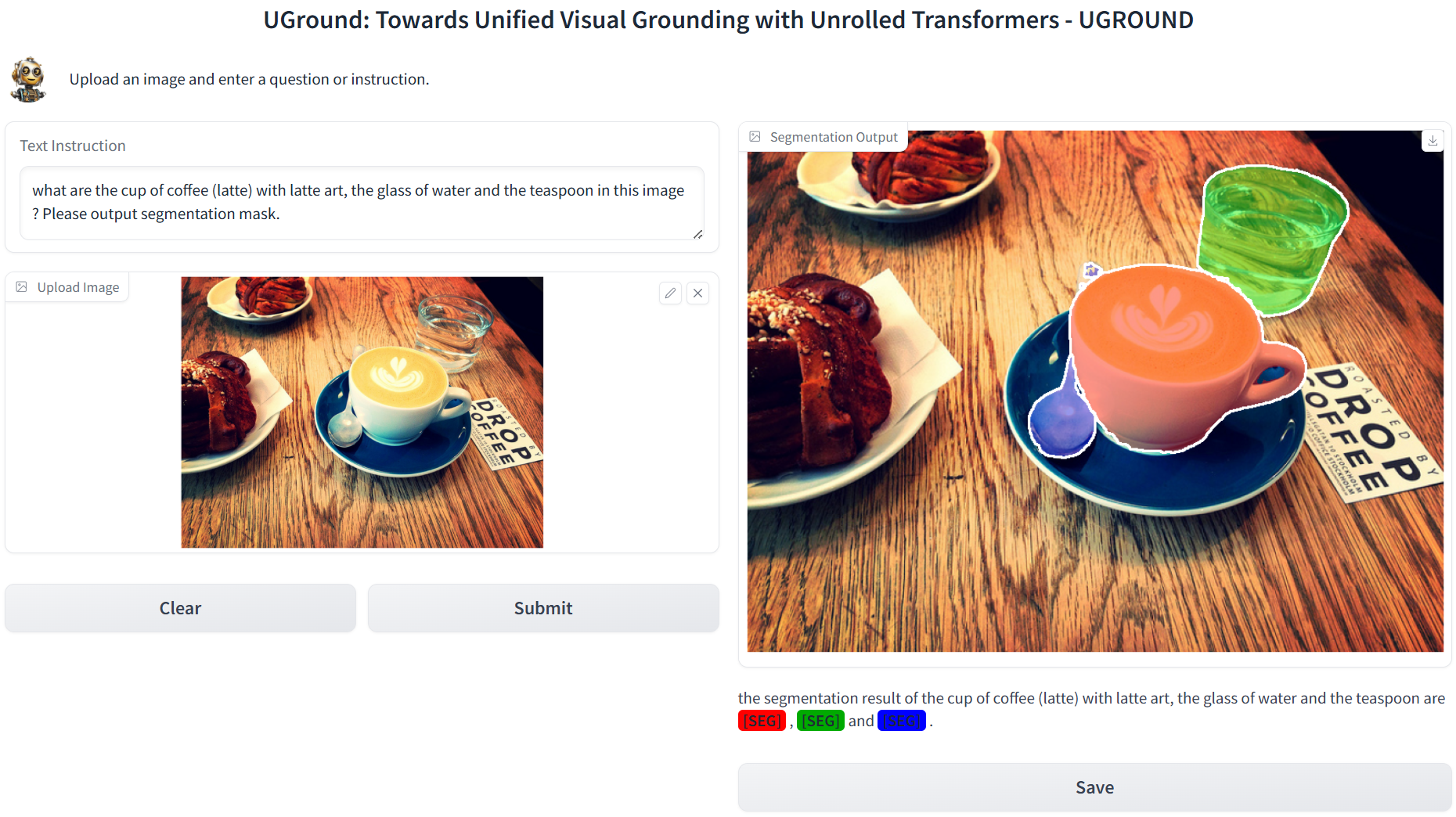}
\end{center}
\caption{The UGround dashboard, demonstrating a unified interface that supports multiple visual grounding models, \ie, LISA~\cite{lai2024lisa}, SESAME~\cite{wu2024see}, PixelLM~\cite{ren2024pixellm}, GSVA~\cite{xia2024gsva}, and READ~\cite{qian2024reasoning}. This example illustrates the result of a single UGround inference.}
\label{fig:vis_multi_target}

\end{figure*}

\textbf{The UGround dashboard.}
As shown in Fig.~\ref{fig:vis_multi_target}, we developed a web-based dashboard to showcase the capabilities of UGround. The system accepts multimodal inputs, specifically a user-uploaded image paired with a natural language instruction. In response, the interface generates two outputs: (1) a segmentation mask that highlights the image region corresponding to the instruction, and (2) a textual explanation that answers the user’s query.

\section{Ethics and Societal Impact}
\label{sup:statements}
\textbf{Use of Large Language Models.}
During the preparation of this manuscript, we used Large Language Models (LLMs) primarily to aid in writing and polishing the text. Specifically, LLMs were employed for grammar correction, spelling improvements, and sentence restructuring to enhance readability and clarity. In terms of technical content, LLMs assisted only in routine tasks such as selecting tensor indices and writing dataset DataLoader code; all algorithmic ideas, methodology, and experimental designs were developed entirely by the authors. All model-generated text was reviewed and edited by the authors, who take full responsibility for the final content of this paper.

\section{Limitations and Future Work}
We observe that UGround generally understands which object is being referred to, which means that the semantic grounding is correct (\ie, the language responses are accurate), yet the predicted masks often fail to capture the full spatial extent of the object. The masks are frequently fragmented, perforated with holes, spatially offset, or spilled into irrelevant regions; in some cases, they even collapse entirely. This indicates that the issue does not stem from semantic misinterpretation, but from imperfect cross-modal geometric alignment.

Although the \texttt{<SEG>} token is intended to represent the “target object,” it is fundamentally a language-space embedding whose feature distribution is not inherently aligned with the spatial structure of visual tokens. In other words,
\texttt{<SEG>} token $\ne$ \text{spatially aligned embedding}, \text{embedding space} $\ne$ \text{image space}.
For example, a distant patch of blue sky and a blue notebook on a desk may appear close in embedding space due to similar color or texture, even though they are far apart in the image. Thus, the \texttt{<SEG>} token captures semantic concepts rather than the spatial statistics of an object’s distribution. While using the mask as a prompt does introduce spatial cues, the internal decoding mechanism of SAM remains largely unexplored, making it unclear how these cues propagate into its actual segmentation behavior.

\begin{figure*}[h]
\begin{center}
\includegraphics[width=0.98\linewidth]{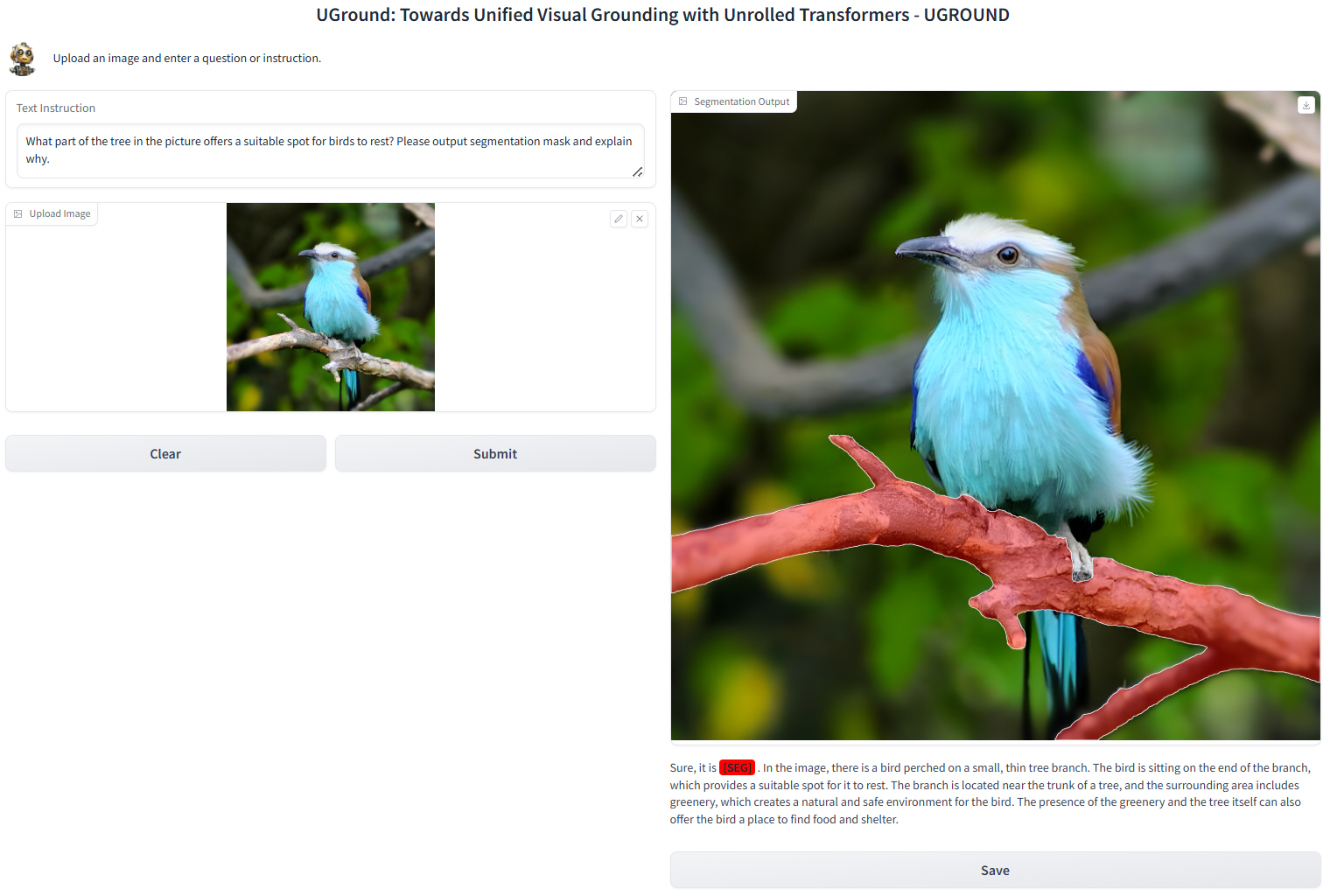}
\end{center}
\label{fig:vis}
\vspace{-0.3cm}

\end{figure*}

\end{document}